\documentclass[12pt]{iopart}
\usepackage{graphicx}
\usepackage{xurl}

\usepackage{xcolor}

\usepackage{float}
\usepackage{natbib}

\usepackage{hyperref}

\begin{document}

\title[]{Wall-to-Wall Amazon Forest Height Mapping with Planet NICFI, Aerial LiDAR, and a U-Net Regression Model}

\author{Fabien H Wagner$^{1,2}$, Ricardo Dalagnol$^{1,2}$, Griffin Carter$^{1}$, Mayumi CM Hirye$^{1,3,4}$, Shivraj Gill$^{1}$, Le Bienfaiteur Sagang Takougoum $^{1,3}$, Samuel Favrichon $^{2,5}$, Michael Keller$^{2,6}$, Jean P. H. B. Ometto $^{7}$, Lorena Alves $^{1,3}$, Cynthia Creze $^{1,3}$, Stephanie P George-Chacon $^{1,3}$, Shuang Li $^{1}$, Zhihua Liu $^{1}$, Adugna Mullissa $^{1,3}$, Yan Yang $^{1}$, Erone G Santos$^{2}$, Sarah R Worden$^{2}$, Martin Brandt$^{1,8}$,  Philippe Ciais$^{1,9}$, Stephen C. Hagen$^{1}$ and Sassan Saatchi$^{1,2,3}$}

\address{$^1$ CTrees, Pasadena, CA 91105, US}
\address{$^2$ Jet Propulsion Laboratory, California Institute of Technology, 4800 Oak Grove, Pasadena, CA 91109, USA}
\address{$^3$ Institute of Environment and Sustainability, University of California, Los Angeles, CA, USA}
\address{$^4$ Quapá Lab, Faculty of Architecture and Urbanism, University of São Paulo, 05508080, São Paulo, SP, Brazil}
\address{$^5$ Gamma Remote Sensing Ag, Gumligen, Switzerland}

\address{$^6$ USDA Forest Service, International Institute of Tropical Forestry, Rio Piedras, Puerto Rico, USA }
\address{$^7$ Remote Sensing Division, National Institute for Space Research---INPE, S\~{a}o Jos\'{e} dos Campos 12227-010, SP, Brazil;}
\address{$^8$ Department of Geosciences and Natural Resource Management, University of Copenhagen, Copenhagen, 1350, Denmark}
\address{$^9$ Laboratoire des Sciences du Climat et de l’Environnement, CEA-CNRS-UVSQ, CE Orme des Merisiers, Gif sur Yvette, 91190, France}

\ead{wagner.h.fabien@gmail.com}

\vspace{10pt}
\begin{indented}
\item[] \today 
\end{indented}

\newpage

\begin{abstract}
Tree canopy height is a key indicator of forest biomass, productivity, and structure, yet measuring it accurately at regional or larger scales, whether from the ground or remotely, remains challenging. The objective of this study is to generate the first complete canopy height map of the Amazon forest at $\sim$4.78 m resolution using Planet NICFI imagery and deep learning. Specifically, we (i) trained a U-Net regression model with canopy height models (CHM) derived from tropical airborne LiDAR and their corresponding Planet NICFI images to estimate canopy height, (ii) evaluated the accuracy of our map against existing global products based on Sentinel-2/1 and Maxar Vivid2 imagery, and (iii) assessed its capacity to capture small-scale canopy height changes. Tree height predictions on the validation sample had a mean absolute error of 3.68 m, with minimal systematic bias across the full range of tree heights in the Amazon forest. The main biases are a slight overestimation (up to 5 m) for heights of 5–15 m and an underestimation for most trees above 50 m. Outperforming existing global model-based canopy height products in this region, the model accurately estimated canopy heights up to 40–50 m with minimal saturation. We determined that the Amazon forest has an average canopy height of $\sim$22 m (standard deviation $\sim$5.3 m) and exhibits large-scale patterns, ranging from the tallest forests of the Guiana Shield to shorter forests along wetlands, rivers, rocky outcrops, savannas, and high elevations. Events such as logging or deforestation could be detected from changes in tree height, and the results demonstrated a first success in monitoring the height of regenerating forests. Finally, the map of the Amazon forest canopy height is displayed.
\end{abstract}

\newpage
\section{Introduction}

The Amazon forest is the most biologically diverse area on the planet, containing around 20\% of the diversity of vascular plants, with the number of tree species alone estimated at around 16,000 species \citep{TerSteege2013,ZapataRios2022}. In terms of area, the Amazon forest represents $>$ 50\% of the world’s remaining tropical rainforests, and much of it is still botanically intact \citep{Hansen850,Hubbell2008}. The Amazon forest biomass constitutes one of the largest terrestrial carbon pools in the world, estimated at $\sim$ 150 to 200 GtC \citep{Mo2023,Gatti2021}. While still functioning as a carbon sink, this role is increasingly challenged by disturbances \citep{Pan2024, Rosan2024}.  Recently, the Amazon rainforest has experienced significant changes in tree cover and forest structure due to extensive deforestation \citep{INPE2024, Prodes2021}, degradation from logging and fire \citep{Matricardi2020, Qin2021, Lapola2023, Bourgoin2024}, and increased drought and extreme conditions \citep{Gloor2013, Marengo2016, Silva2023}. Some forest areas are regenerating after deforestation, but accurately estimating this regrowth and its sink potential remains in its early stages \citep{heinrich2021, SilvaJunior2020, Heinrich2023, Bastin2019, Mo2023}. To protect this ecosystem, support policies to limit deforestation, improve timber-harvesting practices, and identify potential areas for forest restoration requires better forest structure and biomass estimation at individual tree level \citep{Pan2024, Mo2023}. While these metrics can be derived from canopy height \citep{asner2014m, lim2003lidar, Longo2016}, an Essential Biodiversity Variable \citep{skidmore2021p}, domain-scale estimation of tree height remains challenging.

Airborne LiDAR provides the most accurate forest canopy height estimates, but its high cost limits coverage to small or sparse areas. In the Brazilian Amazon, significant efforts have been made to flight airborne LiDAR and capture forest structure diversity through scientifically-based LiDAR sampling design and over permanent plot with field measurements \citep{Ometto2023, DosSantos2022,Csillik2024}. The GEDI \citep{dubayah2020g,Liu2025} and ICESat \citep{Markus2017,Lefsky2005,Lefsky2010} missions provide additional Amazon coverage, but their observations are sparse and have coarser resolution than airborne LiDAR. Current approaches use machine/deep learning with multispectral and radar imagery, trained on LiDAR or GEDI reference data, to estimate tree height over larger areas. These models can infer canopy height from local to global scales using various data sources, including Landsat \citep{potapov2021m}, Sentinel-2 \citep{lang2022, astola2021, lang2019}, combined Sentinel-1/2 imagery \citep{Schwartz2024, Pauls2024,fayad2023vision,Schwartz2023,ge2022}, Planet imagery \citep{Liu2023, csillik2019, huang2022e}, very high-resolution (VHR) airborne data \citep{Wagner2024, Li20233, karatsiolis2021i, li2020h}, and Maxar satellite VHR images \citep{tolan2023sub,illarionova2022e}.

Global canopy height models typically use Landsat or Sentinel-1/2 \citep{potapov2021m, lang2022, Pauls2024}, but their spatial resolutions (30 m or 10 m) are too coarse to identify individual trees. Maxar Vivid2 commercial imagery is also used by Meta \citep{tolan2023sub}, but it suffers from heterogeneity due to system properties and data collection strategy, and incompleteness in the Amazon due to clouds. The complete high-resolution canopy height model for the Amazon is still to be done. Planet NICFI images \citep{Planet2017} currently provide the best data for Amazon monitoring, offering 4.78 m spatial resolution multispectral data (RGB + NIR) between 30$^\circ$N and 30$^\circ$S. Monthly cloud-free composites are created from the best daily acquisitions of the month. Despite limited radiometric accuracy \citep{Pandey2021}, Planet data have been used successfully to map tree cover \citep{Wagner2023,Carter2024} and forest degradation \citep{Dalagnol2023, CTreesREDD+AI2024} in the Amazon. Planet NICFI is currently the only dataset that enables consistent temporal monitoring of canopy structure in the Amazon at less than 10 m spatial resolution and represents a unique resource for advancing Amazon canopy height mapping.

Examples of regional or larger-scale canopy height mapping with deep learning are rare in the tropics and typically come from global datasets \citep{lang2022, tolan2023sub, Pauls2024}, while such methods are slightly more developed in other regions, such as boreal and temperate forests \citep{astola2021, fayad2023vision, Schwartz2023, Schwartz2024, Liu2023, Wagner2024, Li20233, illarionova2022e}. With the exception of Tolan's model \citep{tolan2023sub}, which uses a fine-tuned foundation model, most other studies demonstrate that tree height estimation can be achieved with great accuracy using encoder-decoder deep learning architectures trained locally \citep{Rolf2024}, such as U-Net \citep{Ronneberger2015}. This highlights locally trained U-Nets as the architecture of choice for mapping canopy height in the Amazon.

The objective of this study is to generate the first complete canopy height map of the Amazon at $\sim$4.78 m resolution using Planet NICFI imagery and deep learning. Specifically, we aim to (i) demonstrate that U-Net architectures trained with canopy height models (CHM) derived from tropical airborne LiDAR, together with their time-aligned Planet NICFI images, can provide reliable canopy height estimates at $\sim$4.78 m resolution, (ii) evaluate the performance of our map against existing global products, including those from Sentinel-2 and -1 (10 m) and Maxar Vivid2 mosaic imagery (0.5 m), and (iii) assess its ability to detect small-scale changes in canopy height resulting from deforestation, logging, and regeneration. Finally, we present the Amazon forest tree height map for the 2020–2024 period.

\section{Methods}

\begin{figure}[ht]
\centering
\includegraphics[width=1\linewidth]{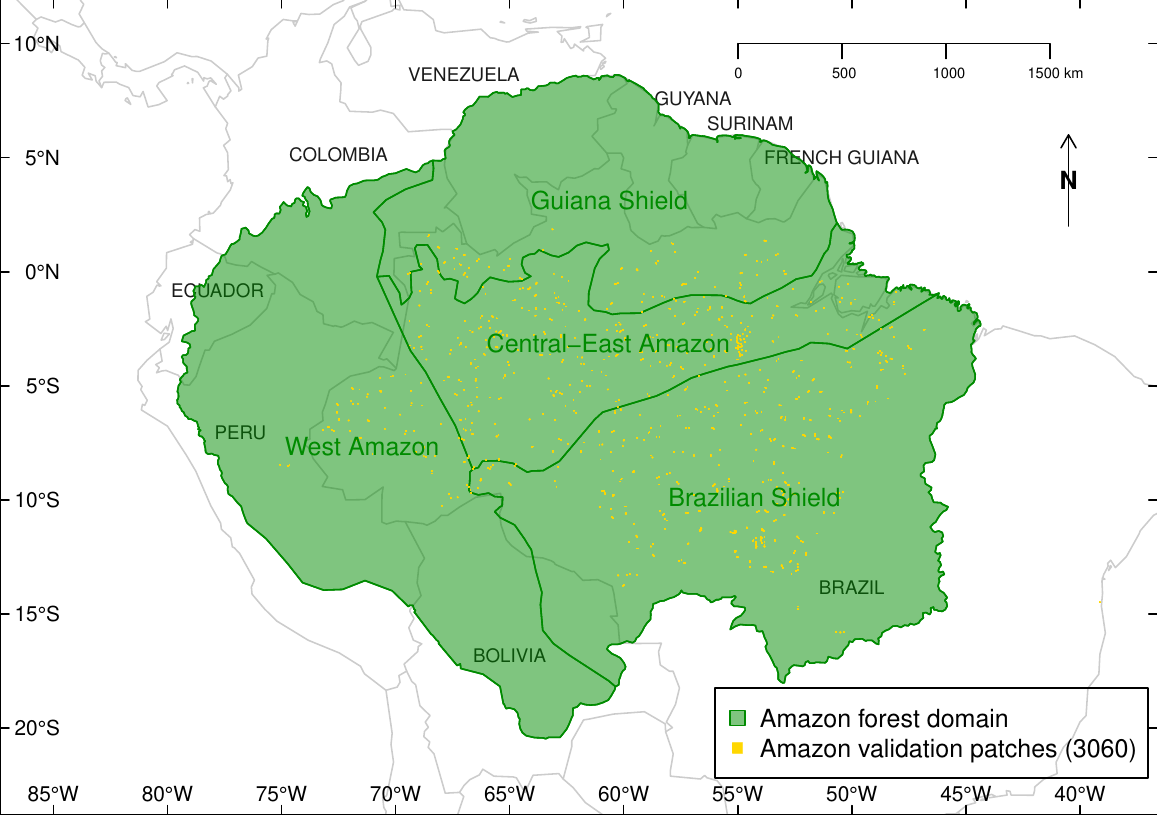}
\caption{Extent of the Amazon forest domain used in this study and its partition into four regions according to \citet{Feldpausch2011}. The location of the 3060 validation patches of 256 $\times$ 256 pixels in the Amazon domain is shown in gold. The validation locations approximately represent the locations of all LiDAR flights in the Amazon used in the study, as at least one patch per flight was used for validation. The additional 376 validation patches in the Atlantic forest domain, mostly located over São Paulo city, are not displayed on the map.}
\label{FigMeteMet1}
\end{figure}

\subsection{Data} 
\subsubsection{Planet NICFI satellite images}\label{satimg}

To train the model and predict the tree canopy height of the Amazon forest, we used Planet satellite images from NICFI (\url{https://www.nicfi.no/}) \citep{Planet2017} at 4.78 m resolution. Approximately 1,240,000 images covering the Amazon (8,260,920 km$^2$, Fig. \ref{FigMeteMet1}) were downloaded through Planet API, with biannual coverage (2015-2020, 10 dates) and monthly coverage (2020-2024, 46 dates). The Planet NICFI images are distributed in tiles of size 4096 $\times$ 4096 pixels ($\sim$20 $\times$ 20 km), and in the following text, \emph{tile} refers to the Planet NICFI tile. The 12-bit Red (0.650-0.682 $\mu$m), Green (0.547-0.585 $\mu$m), Blue (0.464-0.517 $\mu$m), and NIR (0.846-0.888 $\mu$m) bands were first scaled 0-2540 (RGB truncated to 0-2540 and NIR divided by 3.937), scaled to 8-bit (0-255), and combined into RGBNIR composites without atmospheric correction.

\subsubsection{LiDAR airborne datasets}

The Embrapa LiDAR datasets were collected under the "Sustainable Landscapes - Brazil" Program to assess carbon stocks and human impacts in the Amazon Forest \citep{DosSantos2022}. LiDAR data was acquired across Amazon, Cerrado, and Atlantic forest from 2008-2018, with ongoing collection (\url{https://www.paisagenslidar.cnptia.embrapa.br/}). We only kept LiDAR data collected in 2015 or later to overlap with the Planet NICFI dates. The publicly available LAZ files have $\sim$10 points/m$^2$ resolution and each covers 1~km$^2$. We used 1,149 LAZ files of 70 flights from 23 Brazilian sites (2015-2018) and 1 Peruvian site (2017).

The EBA project collected LiDAR transects across the Brazilian Amazon (2016-2018) \citep{Ometto2023}, available at \url{https://zenodo.org/records/7636454}. Each Transect (713 total) covered 375 hectares minimum using a Trimble Harrier 68i sensor with 4 returns/m$^2$, 30$^{\circ}$ field of view, and 600 m flight height. Data accuracy was $<$1m horizontal and $<$0.5m vertical. We used 19,302 LAZ files ($\sim$1000 $\times$ 1000 m each).

The São Paulo Metropolitan Region dataset was collected in 2017 and is available at \url{https://registry.opendata.aws/pmsp-lidar/}. This data was used to train the model on diverse backgrounds (pasture, soil, buildings) and Atlantic forest areas. The survey used an Optech Gemini sensor with 100-125 KHz frequency, 18-25$^{\circ}$ scan angle, 0.25 mrad beam divergence, 700 m flight height, and 8 returns/m$^2$ density. We used 5,554 LAZ files (500 $\times$ 500 m each).

LiDAR point clouds (.LAZ format) were processed into canopy height models (CHM) using the \texttt{LidR} R package \citep{Roussel2021,Roussel2020}. First, outliers in the LiDAR point clouds were eliminated through denoising using the $ivf$ algorithm, with a resolution set to 1 m and a maximum of 5 surrounding points \citep{Roussel2021}. Second, the point cloud was classified using the progressive morphological filter (pmf) algorithm with parameters ws = seq(4, 42, 3), or ws = seq(3, 12, 3) if the algorithm had difficulty converging, and th = seq(0.1, 1.5, length.out = length(ws)). Third, the digital terrain model (DTM) and digital surface model (DSM) were generated at a spatial resolution of 1 m. The $TIN$ algorithm \citep{Roussel2021} was applied to compute the DTM, while the DSM was generated using the $pitfree$ algorithm with thresholds set to [0, 2, 5, 10, 15] and a maximum edge length of [0, 1.5]. All the parameters were chosen to produce a smooth 1 m canopy height model without any holes reaching the ground inside the crowns, which can occur with other parameterizations and could impede a good fit of the U-Net model, as the model focuses on these large, unpredictable errors. The final CHM was created by subtracting DTM from DSM, scaling by 2.5, and converting to 8-bit integer format to optimize height resolution within the 0-255 range.

 \subsubsection{Building footprints masking}

Google's Building Footprints for South America (\url{https://sites.research.google/open-buildings/}) was used to remove building heights from CHM. Building polygons were buffered (5m outside cities, 15m inside) and rasterized to match CHM resolution. All CHM pixels intersecting these buffered footprints were set to zero.

\subsubsection{Global tree height datasets}

We evaluated our model against three deep learning-based global canopy height datasets : Meta's 0.5 m resolution map using DINOv2 and Maxar RGB imagery \citep{metachm2023,tolan2023sub}, Lang's 10 m resolution map using CNN and Sentinel-2 data \citep{lang2022}, and Pauls's 10 m resolution map using CNN with Sentinel-1/2 data \citep{Pauls2024}. All used GEDI or aerial LiDAR as reference data.

\subsubsection{Deforestation and logging datasets}

To analyze tree canopy height changes, we used two Planet NICFI-based datasets: a deforestation dataset showing monthly tree cover changes \citep{Wagner2023,CTreesREDD+AI2024} and a logging dataset showing bi-annual activities \citep{Dalagnol2023,CTreesREDD+AI2024} in pan-tropical evergreen forests. These datasets were used to find areas of deforestation, regeneration, and selective logging in the Amazon and their date of occurrence.

\subsection{Model}
\subsubsection{Neural Network Architecture}

\begin{figure}[ht!]
\centering
\includegraphics[width=0.80\linewidth]{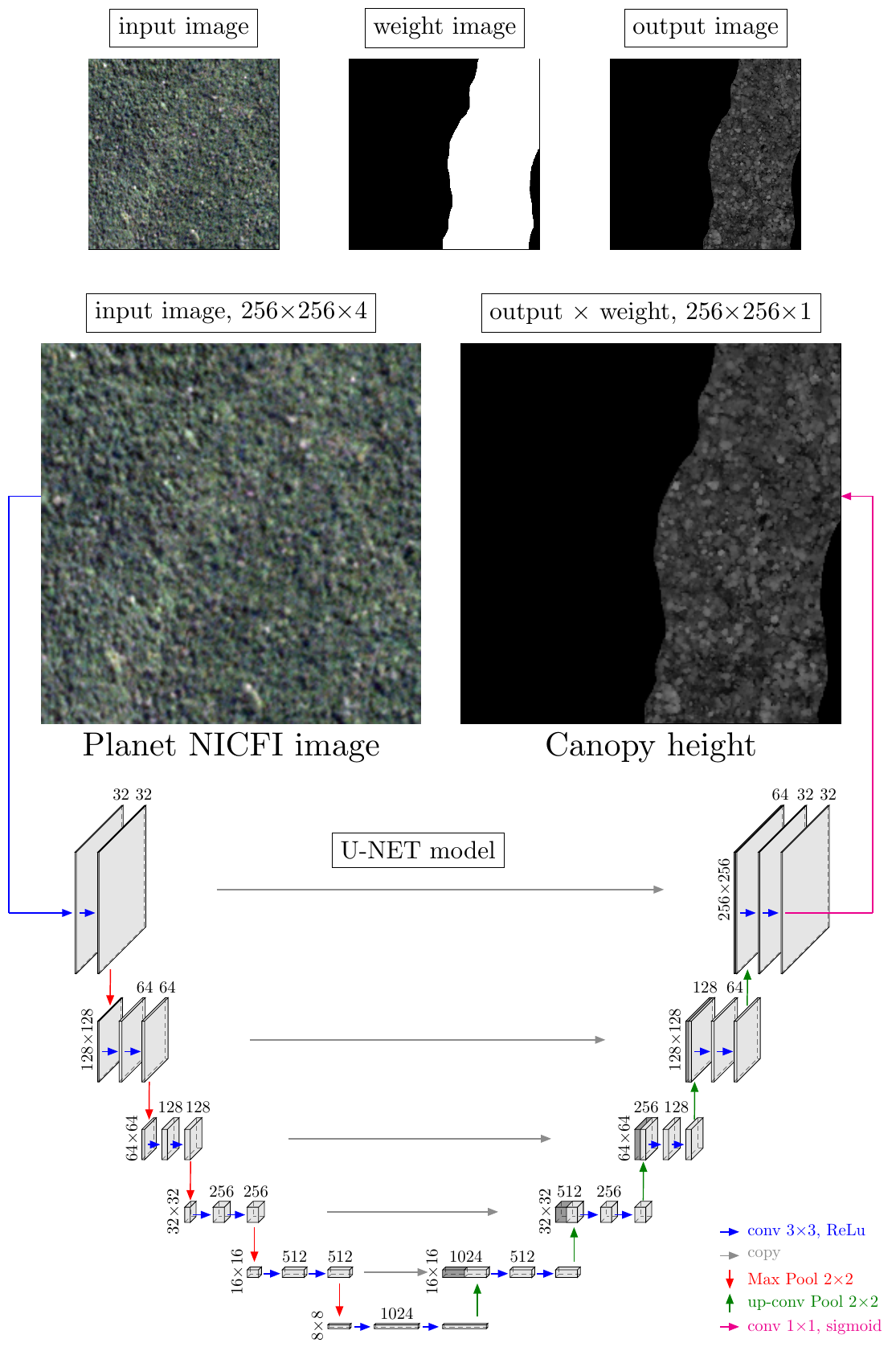}
 \caption{U-Net architecture used to estimate canopy height from Planet NICFI imagery, adapted from \citet{Ronneberger2015}. The number of channels is shown above each cuboids, and the vertical dimensions represent the row and column sizes in pixels. Operations such as convolutions, skip connections, max pooling, and upsampling, along with their respective sizes, are illustrated with colored arrows. For the weight, the white area (value = 1) corresponds to pixels with CHM data, while the black area (value = 0) represents NA values.}
 \label{FigUnet}
 \end{figure}

We used a U-Net model \citep{Ronneberger2015} to estimate Amazon forest canopy height from Planet NICFI images, Fig. \ref{FigUnet}. The model processes 256 $\times$ 256 pixel RGB-NIR inputs (4.78m resolution) to produce height predictions at pixel level. Implementation used R with Keras and TensorFlow 2.10 \citep{CoreTeam2016, chollet2015keras,AllaireChollet,allaireTang,AbadiAgarwalBarhamEtAl2015}.

\subsubsection{Training and validation}

To generate samples for model training and validation, for a given LiDAR flight, we selected all Planet NICFI tiles that overlapped its extent and were closest in date to the flight. Then, to increase sample size and reduce cloud effects, we also included Planet NICFI images from adjacent dates, assuming no significant forest height change within up to one year. This corresponds to $\pm 6$ months for the biannual Planet NICFI time series. The 1 m CHM data was resampled to 4.78 m Planet resolution using median values, along with a binary mask indicating CHM data presence. The median was used to minimize the impact of outliers in the 1 m CHM. Both Planet images and resampled CHM data (4096 $\times$ 4096 pixels) were split into 256 $\times$ 256 patches using \texttt{gdal\_retile} \citep{gdal2019}. To maximize the use of the LiDAR dataset and ensure independent validation, we selected a 256×256 sub-image for validation from each Planet tile intersecting the flight lines, with the remaining data used for training. The validation data came from the same flights as the training data but did not spatially overlap with the training samples. The dataset included 50,724 patches of 256 $\times$ 256 pixels (47,288 training, 3,436 validation), with validation patches distributed randomly along the flight lines across Amazon (3,060, as shown in Fig. \ref{FigMeteMet1})  and Atlantic (376) forests.

The U-Net model was trained using SGD optimization with Adam optimizer (learning rate 0.0001), weighted mean squared error loss, horizontal and vertical flips for data augmentation, and 1,000 epochs with 32-image batches. The best model achieved 0.0008285521 validation loss and 3.62 m mean absolute error, training in under 24 hours on an Nvidia RTX4090 GPU.

For the 3436 validation images, we assessed the accuracy of the predicted canopy heights at the pixel level using the Root Mean Square Error (RMSE), the Mean Absolute Error (MAE), the Bias, along with their percentages relative to the mean reference canopy height, RMSE\%, MAE\%, and Bias\%, eqn. \ref{eqnid0}-\ref{eqnid6}. All accuracy metrics were also computed using the three global canopy height datasets on the validation images to enable comparison with our results.

\begin{eqnarray}
\overline{H}_{\mathrm{ref}} &=& \frac{1}{N} \sum_{i=1}^{N} H_{\mathrm{ref}, i} \quad \mathrm{(mean\,of\,reference\,heights)} \label{eqnid0}\\
RMSE &=& \sqrt{ \frac{1}{N} \sum_{i=1}^{N} \left( H_{\mathrm{pred}, i} - H_{\mathrm{ref}, i} \right)^2 } \label{eqnid1} \\
RMSE\% &=& \frac{RMSE}{ \overline{H}_{\mathrm{ref}} } \times 100 \label{eqnid2}\\
MAE &=& \frac{1}{N} \sum_{i=1}^{N} \left| H_{\mathrm{pred}, i} - H_{\mathrm{ref}, i} \right| \label{eqnid3}\\
MAE\% &=& \frac{MAE}{ \overline{H}_{\mathrm{ref}} } \times 100 \label{eqnid4}\\
Bias &=& \frac{1}{N} \sum_{i=1}^{N} \left( H_{\mathrm{pred}, i} - H_{\mathrm{ref}, i} \right) \label{eqnid5}\\
Bias\% &=& \frac{Bias}{ \overline{H}_{\mathrm{ref}} } \times 100 \label{eqnid6}\\
where && \nonumber \\
N &=& \mathrm{number\,of\,validation\,samples} \nonumber\\
H_{\mathrm{pred}, i} &=& \mathrm{predicted\,canopy\,height\,at\,pixel\,} i \nonumber\\
H_{\mathrm{ref}, i} &=& \mathrm{reference\,canopy\,height\,at\,pixel\,} i \nonumber
\end{eqnarray}

Finally, median and percentiles (0.05-0.95) of canopy height were computed within each validation area to summarize the canopy height distribution and to provide complementary insight into model performance across the distribution of observed values.
 
\subsection{Amazon Forest 2020–2024 Canopy Height Map Production}
\label{masks}

For prediction, NICFI tiles were padded to 4224 $\times$ 4224 with adjacent tiles pixels values, then trimmed by 64 pixels to reduce border artifacts \citep{Ronneberger2015}. Mean Amazon forest height (2020-2024) was calculated at pixel level from observations between 2016-2024, excluding clouds, cloud shade, and areas deforested before 2020 or regenerated after 2020 \citep{Wagner2023,CTreesREDD+AI2024}. Pixels with forest loss between 2020 and 2024 were retained in the analysis which may lower mean canopy height mainly near Amazon forest edges where deforestation occurred.
Cloud shade in the monthly height mask appears as continuous areas with very high values that do not correspond to real forest structure. We generate a cloud shade mask to identify and remove these false high-value areas as follows. First, we apply thresholding to create a binary mask of pixels with height values greater than 32 meters. Next, we apply erosion using a 16 $\times$ 16 kernel to eliminate isolated high pixels, such as single crowns or noise. Then, we apply dilation with a 512 $\times$ 512 kernel to expand the remaining high-value areas. Finally, we use the resulting mask to exclude cloud shade pixels from the mean height computation. If no valid values for a pixel are available across the entire time series, which occurs only in regions of extreme shading or in areas with very tall trees, we gradually increase the threshold to 36 meters and, if necessary, to 40 meters. For the forested pixels, height $\geq$5 m following the definition of \cite{FAO2020}, we computed the mean canopy height using the mean height per tile and the number of valid tree height observations per tile as weights, eqn \ref{eqn1}, where $i$ is a Planet tile, and a valid observation is a pixel with tree height $\geq$ 5 m and more than 5 non-cloud observations. The weighted percentiles 2.5, 5, 25, 50, 75, 95, and 97.5 over the Amazon were also computed using the same per-tile logic. In sections~\ref{cloudsvalid}, \ref{AmzStat}, and \ref{CompaAmz}, we use the Planet NICFI tile as the unit for computing and displaying descriptive statistics.

\begin{eqnarray}
 weighted\,mean & = & \frac{\sum_{i=1}^{n} \left(mean_{i} \times valid\,observations_i\right)}{\sum_{i=1}^{n} valid\,observations_i} 
 \label{eqn1}
 \end{eqnarray}

\section{Results}

\subsection{Accuracy and comparison with global CHM models}

\begin{figure}[!ht]
\centering
 \includegraphics[width=0.75\linewidth]
 {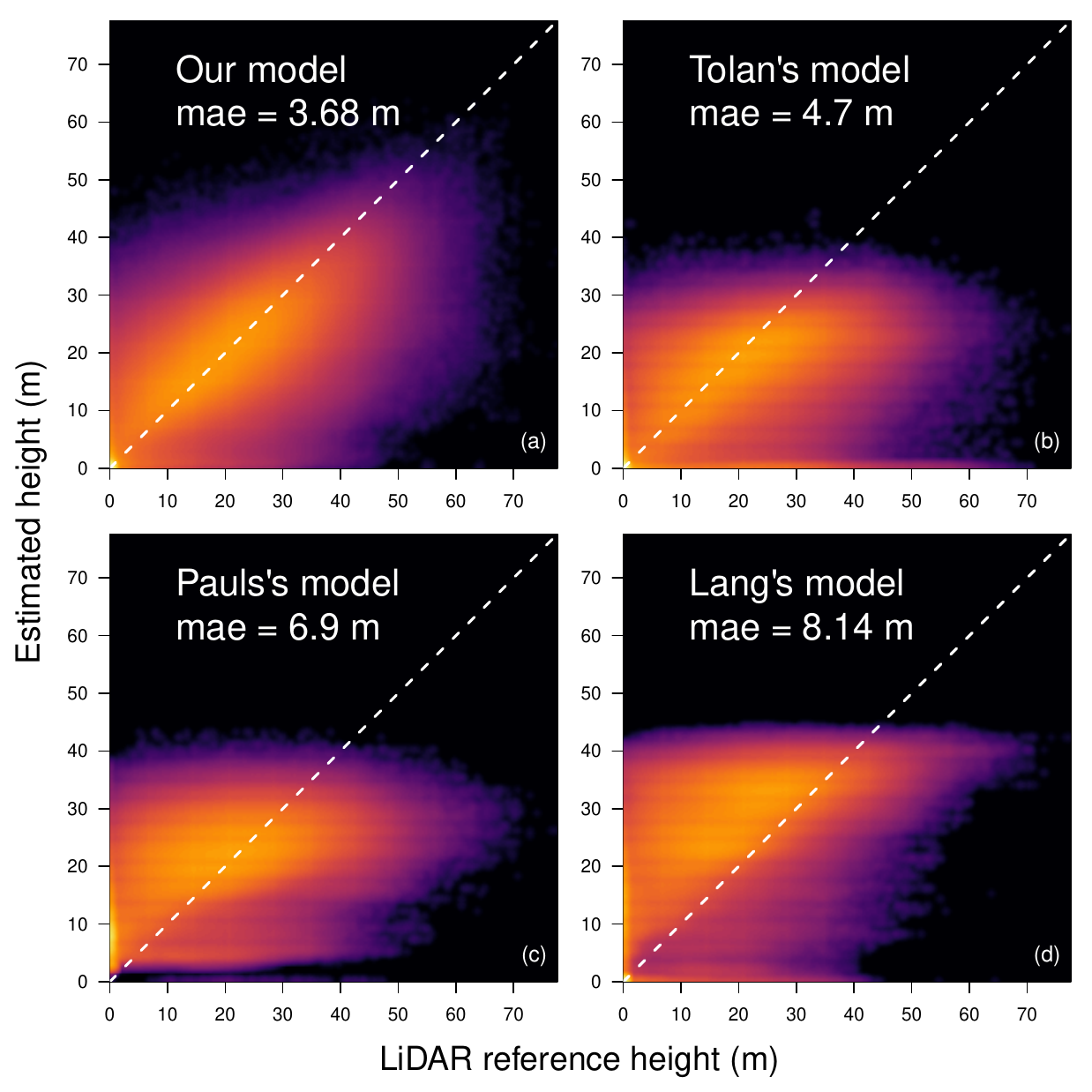}
 \includegraphics[width=0.180\linewidth,trim={0.1cm 8cm 6.0cm 0.1cm},clip]{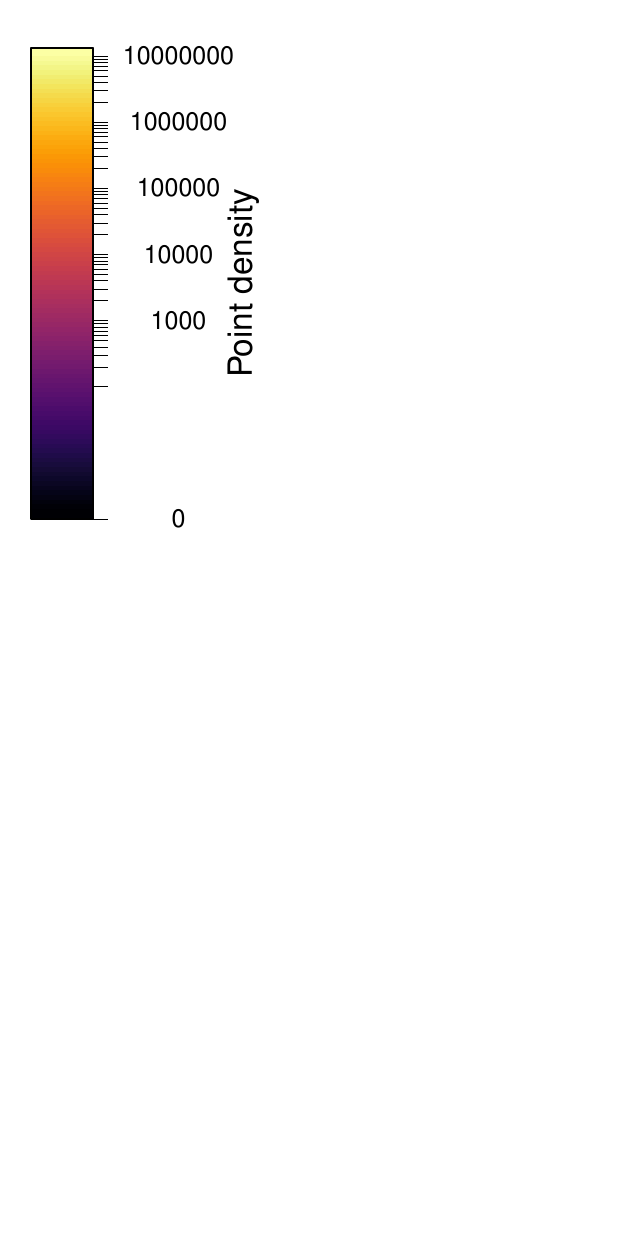}
 \caption{Comparison of predicted versus observed height (m) for the 3,436 validation patches represented as density scatterplots for our canopy height model (a), Tolan's model (b), and Lang's model (c). Tolan's model and Lang's model, with native spatial resolutions of 0.5 m and 10 m respectively, were warped to our 4.78 m spatial resolution using the median and nearest neighbor algorithms, respectively. Each plot contains approximately 63 million points. The 1:1 line and the mean absolute error are depicted in white.} 
\label{FigRes1}
\end{figure}

Our model accurately predicted canopy height, with predictions closely following the 1:1 line across 3,436 validation patches (Fig. \ref{FigRes1}a). The accuracy metrics are ranked consistently with the MAE, Table \ref{tab0}, with our model showing the best performance, followed by Tolan’s model, Pauls’s model, and Lang’s model. Specifically, our model showed an MAE of 3.68 m (MAE\% = 25.87\%), an RMSE of 5.54 m (RMSE\% = 38.91\%), and Bias of -0.28 m (Bias\% = -1.97\%). The predicted height is within 1 m, 3 m and 5 m of the reference in 33.9\%, 55.6\% and 71.6\% of cases, respectively. The model performs well up to 45-50 m but underestimates taller trees, though trees above 50 m represent only 0.06\% of validation pixels. While accurate for non-forest areas, it slightly overestimates heights below 25 m. Note that Planet and CHM images are not co-registered, and NICFI basemaps combine images with varying dates, sensor calibrations, illuminations or acquisition geometries.


\begin{table}[ht]
\caption{Accuracy metrics for the validation dataset for our canopy height model and the three global models of Tolan, Pauls, and Lang. MAE, RMSE, and Bias are reported, along with their percentages relative to the mean reference canopy height.}
\centering
\begin{tabular}{rrrrr}
  \hline
 & Our model & Tolan's model & Pauls's model & Lang's model \\ 
  \hline
MAE (m) & 3.68 & 4.70 & 6.90 & 8.14 \\ 
  MAE\% (\%) & 25.87 & 33.05 & 48.47 & 57.20 \\ 
  RMSE (m) & 5.54 & 7.23 & 8.31 & 10.57 \\ 
  RMSE\% (\%) & 38.91 & 50.82 & 58.36 & 74.28 \\ 
  Bias (m) & -0.28 & -2.30 & 3.54 & 7.41 \\ 
  Bias\% (\%) & -1.97 & -16.16 & 24.85 & 52.10 \\ 
   \hline
\label{tab0}
\end{tabular}
\end{table}

Tolan's 0.5 m Maxar-based model \citep{tolan2023sub} shows good accuracy (MAE: 4.7 m) but saturates at 30-35 m and exhibits cloud-related zero height errors (Fig. \ref{FigRes1}b). Like our model, it overestimates heights below 25 m. Pauls's 10 m Sentinel-1/2-based model \citep{Pauls2024} has 6.9 m MAE and saturates at 40 m (Fig. \ref{FigRes1}c). It overestimates low vegetation height with minimum predictions of 2-3 m. Lang's 10 m Sentinel-2-based model \citep{lang2022} reaches 40-45 m with 8.14 m MAE (Fig. \ref{FigRes1}d). Despite overestimation due to 10 m resolution and training methodology favoring taller canopies, it effectively captures tall trees.

\subsection{Predicted height distribution analysis}

\begin{figure}[!ht]
\centering
\includegraphics[width=0.70\linewidth]{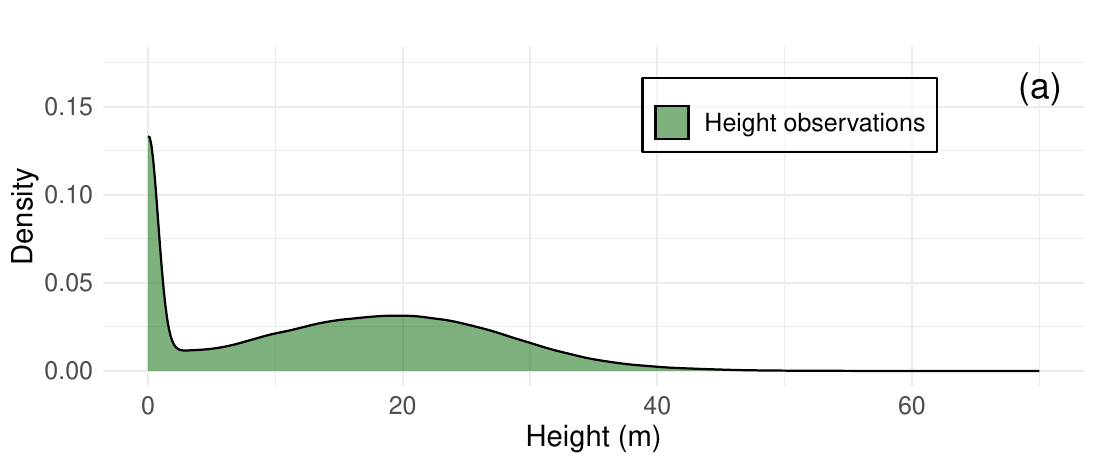}
\includegraphics[width=0.70\linewidth]{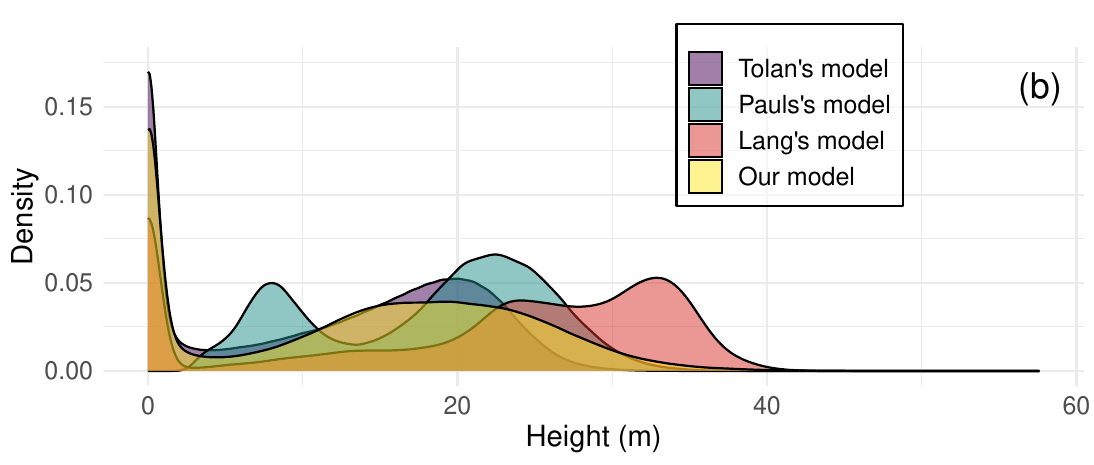}
\includegraphics[width=0.70\linewidth]{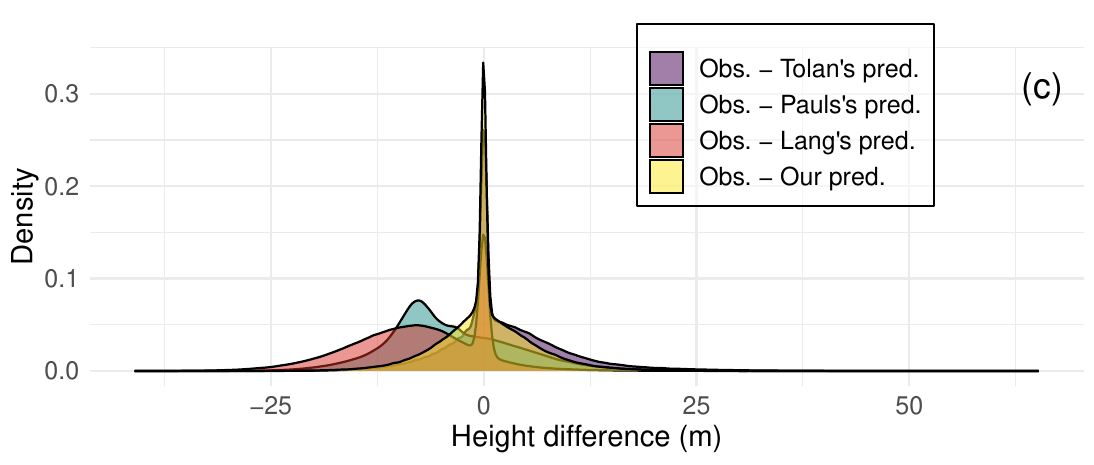}
 \caption{Distribution of observed height in the validation sample (a); predicted height distribution for our model, Tolan's, Pauls's, and Lang's models (b); and differences in predicted height between our model and the others (c). Tolan's, Pauls's, and Lang's models (native resolutions: 0.5 m, 10 m, and 10 m) were warped to our 4.78 m resolution using median or nearest neighbor algorithms.} 
\label{FigRes5}
\end{figure}

Observed validation data shows a bimodal height distribution with peaks at zero and $\sim$20 m (Fig. \ref{FigRes5}a). Our model closely matches this pattern and predicts heights more uniformly than other models (Fig. \ref{FigRes5}b), though slightly underestimating above 20 m and overestimating below.
Tolan's model shows a narrow distribution peaking at 20 m, underestimating taller heights, with more instances of zero height than observed, likely due to cloud cover. Pauls's model exhibits peaks at 22-23 m and 5-10 m, overestimating canopy and low vegetation heights with no near-zero values. Lang's model shows three peaks (near zero, 24 m, and 33 m), generally overestimating heights.

Height difference distributions (Fig. \ref{FigRes5}c) show our model performs best, with Tolan's and Lang's models also peaking at zero. Our predictions are uniformly distributed around zero, while Tolan's model shows excess positive differences due to underestimation above 20-30 m. Both Pauls's and Lang's models exhibit negative difference distributions, indicating systematic height overestimation.

\begin{figure}[!ht]
\centering
 \includegraphics[width=0.70\linewidth]{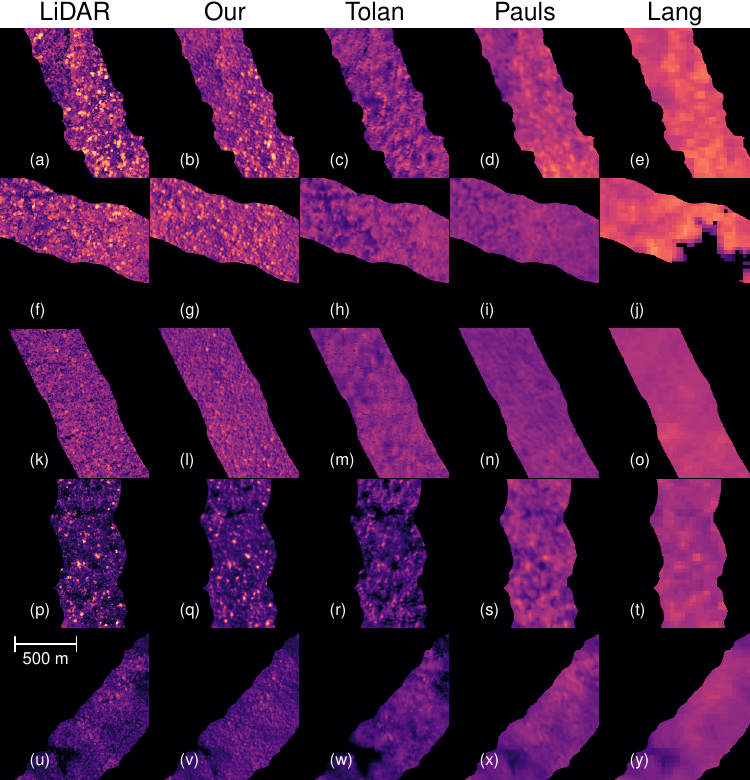}
 \includegraphics[width=0.150\linewidth,trim={0.1cm 0.1cm 0.8cm 0.1cm},clip] {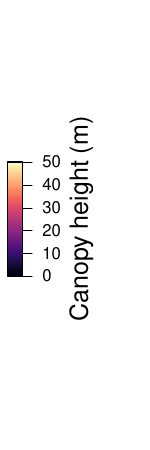}
 \caption{Example of canopy height models observed in the validation dataset (column 1), predicted from our model (column 2), Tolan's model (column 3), and Lang's model (column 4). Tolan's, Pauls's, and Lang's models (native resolutions: 0.5 m, 10 m, and 10 m) were warped to our 4.78 m resolution using the median or nearest neighbor algorithms. Image centroids per row in decimal latitude and longitude: [-7.411495, -53.552856]; [-0.2142329, -62.7154541]; [-11.82972, -53.95935]; [-7.182651, -51.597290]; and [-9.941798, -52.739868].}
\label{FigRes2a}
\end{figure}

\subsection{Example of predictions from the models}

Our model effectively reproduces crown-level CHM across different forest structures (Fig. \ref{FigRes2a}). In dense, tall forests (Fig. \ref{FigRes2a}a-e), we clearly identify crowns despite some height underestimation, while Tolan's model appears blurrier, Pauls's model captures large crowns but struggles with lower points, and Lang's model identifies tree groups well. In uniform dense forests (Fig. \ref{FigRes2a}f-j), our model maintains crown detection, while Tolan's model only captures large trees, Pauls's underestimates tall trees, and Lang's shows cloud-related data gaps. For lower-density forests (Fig. \ref{FigRes2a}k-o), we capture crowns but miss gaps, Tolan's appears blurry, Pauls's matches our heights but blurrier, and Lang's follows overall canopy structure. In forests with isolated large crowns (Fig. \ref{FigRes2a}p-t), our model is the best at crown and gap detection, Tolan's captures large crowns and gaps well, Pauls's overestimates low canopy but detects large crowns, and Lang's captures tall tree groups but overestimates overall. In uniform low forests (Fig. \ref{FigRes2a}u-y), we provide best resolution but may miss lower points that Tolan's captures, while Pauls's overestimates heights except for large crowns, and Lang's captures spatial patterns despite overestimation.

\begin{figure}[!ht]
\centering
 \includegraphics[width=0.70\linewidth]
 {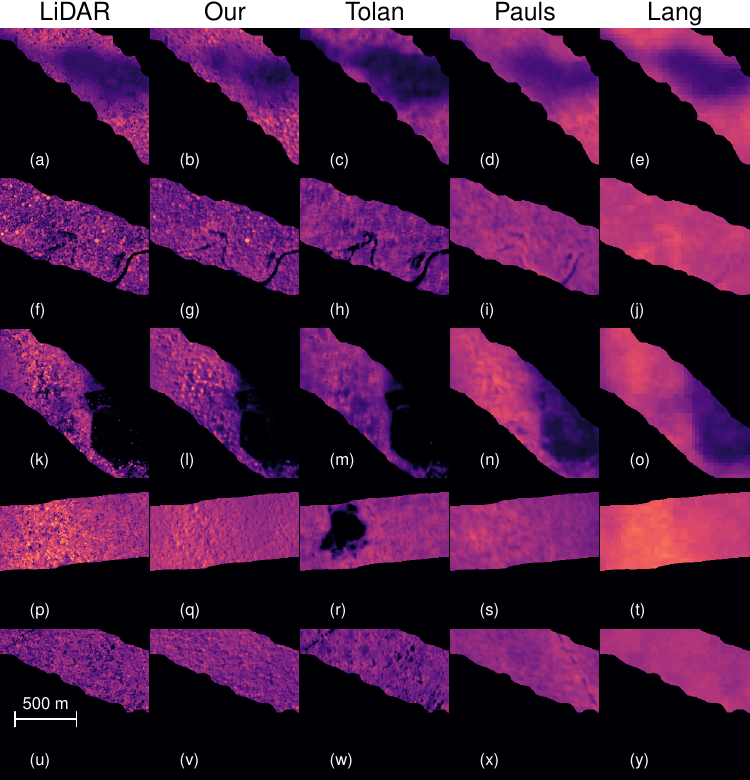}
 \includegraphics[width=0.150\linewidth,trim={0.1cm 0.1cm 0.8cm 0.1cm},clip] {Fig_example_our_lang_pauls_meta_good_to_discuss_legend.pdf}
 \caption{Example of canopy height models observed in the validation dataset (column 1), predicted from our model (column 2), Tolan's model (column 3), and Lang's model (column 4). Tolan's, Pauls's, and Lang's models (native resolutions: 0.5 m, 10 m, and 10 m) were warped to our 4.78 m resolution using the median or nearest neighbor algorithms. Image centroids per row in decimal latitude and longitude: [-10.47161, -56.68396]; [-6.320758, -50.938110]; [-10.50402, -56.64001]; [-4.778995, -61.814575]; and [-10.48241, -53.77258].}
\label{FigRes2b}
\end{figure}

In forests with a smooth and large canopy height transitions (Fig. \ref{FigRes2b}a-e), our model and Tolan's achieve good resolution and low heights, while Pauls's and Lang's models overestimate heights. For rivers inside forest  (Fig. \ref{FigRes2b}f-j), our model and Tolan's correctly map water as zero height, Pauls's only detects larger rivers, and Lang's sometimes misinterprets rivers as tall vegetation. At forest-clearcut boundaries (Fig. \ref{FigRes2b}k-o), our model captures borders and isolated trees well, Tolan's appears blurrier, while Pauls's and Lang's fail to reach zero in cleared areas. In smooth height transitions (Fig. \ref{FigRes2b}p-t), all models perform well, though Tolan's shows cloud artifacts. For small gaps and roads (Fig. \ref{FigRes2b}u-y), Tolan's model detects low points, while our model struggles, and Pauls's and Lang's miss these fine-scale features, though Pauls's captures newer roads (build after our validation sample image was taken in 2018).

\subsection{Height distribution at validation patch-level from our model}

\begin{figure}[!ht]
\centering
 \includegraphics[width=0.95\linewidth]{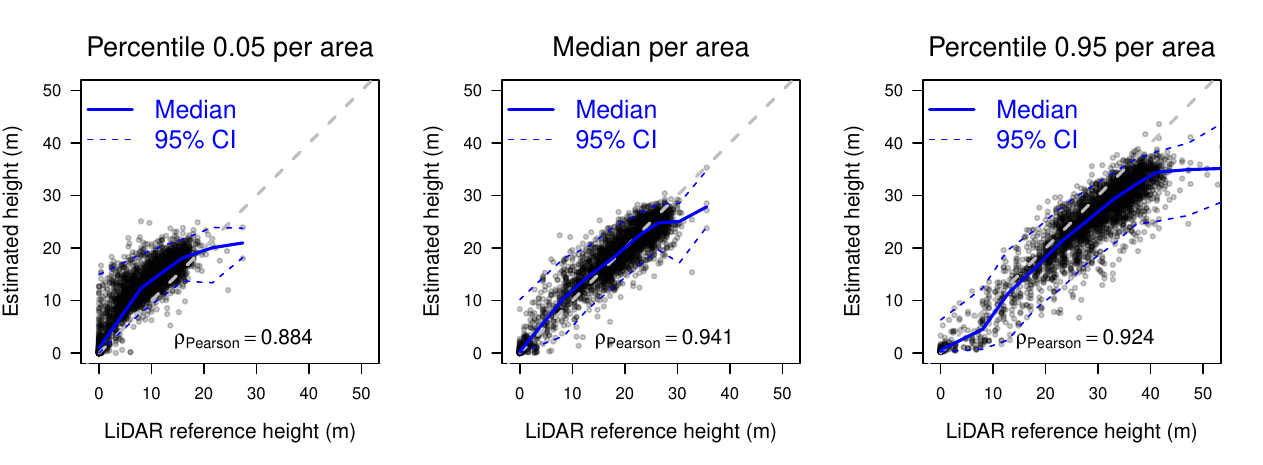}

 \caption{Percentile 0.05 (a), median (b), and percentile 0.95 (c) of heights computed from canopy heights observed and predicted by our model in the 3,436 validation patches. Median of the points and 95\% confidence interval computed by intervals of 5 m is given in blue. Each point represents one validation patch.}
\label{FigRes3}
\end{figure}

Across the 3436 validation patches (Fig. \ref{FigRes3}a), the 5th percentile shows high correlation with LiDAR reference but overestimates heights above 1-2 m, especially at 5-15 m. The median (50th percentile, Fig. \ref{FigRes3}b) demonstrates best accuracy ($\rho$ = 0.927) with predictions well-distributed along the 1:1 line. The 95th percentile (Fig. \ref{FigRes3}c) shows increasing underestimation with height (-5 to -10 m for tallest trees), possibly because taller trees don't necessarily have proportionally larger crowns.

\subsection{Logging detection from height changes}

\begin{figure}[!ht]
\centering
\includegraphics[width=0.95\linewidth, trim={0.2cm 0.2cm 4cm 0.2cm},clip]{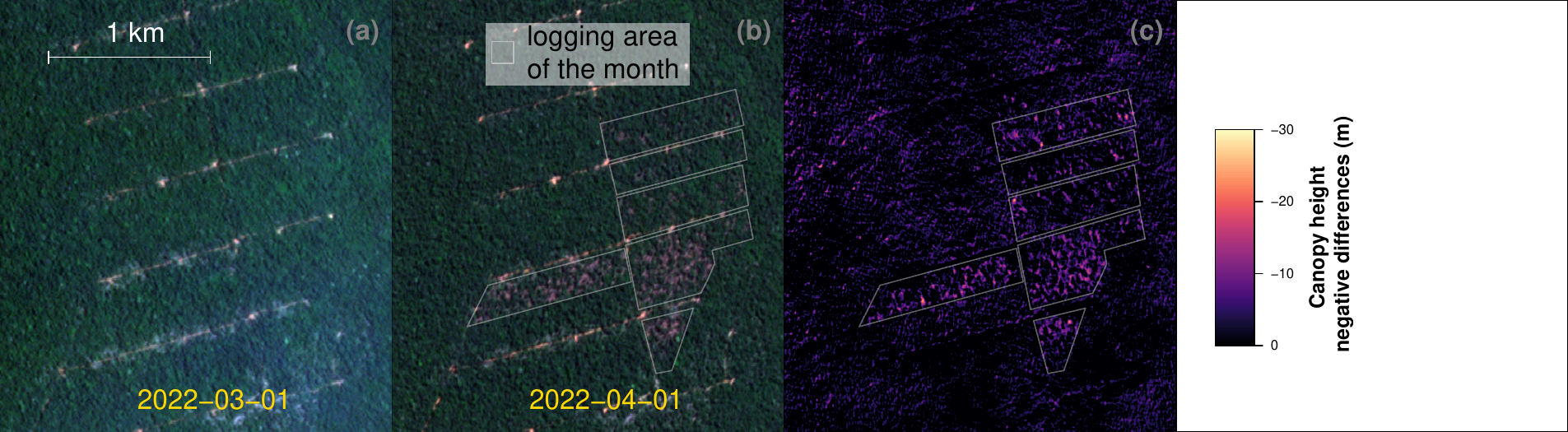}
\includegraphics[width=0.95\linewidth, trim={0.2cm 0.2cm 4cm 0.2cm},clip]{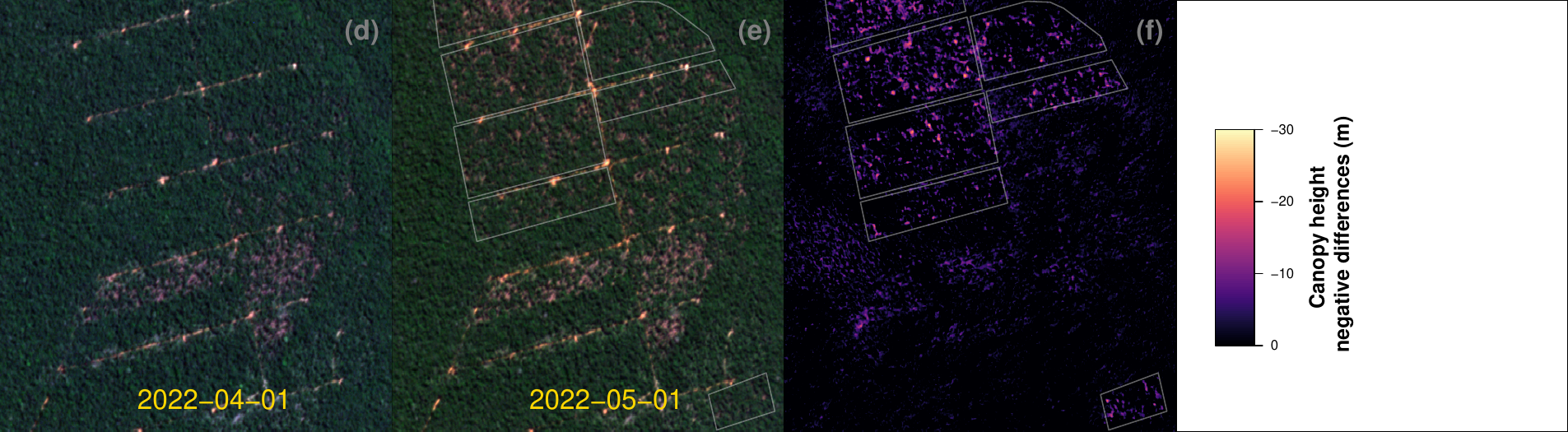}
\includegraphics[width=0.95\linewidth, trim={0.2cm 0.2cm 4cm 0.2cm},clip]{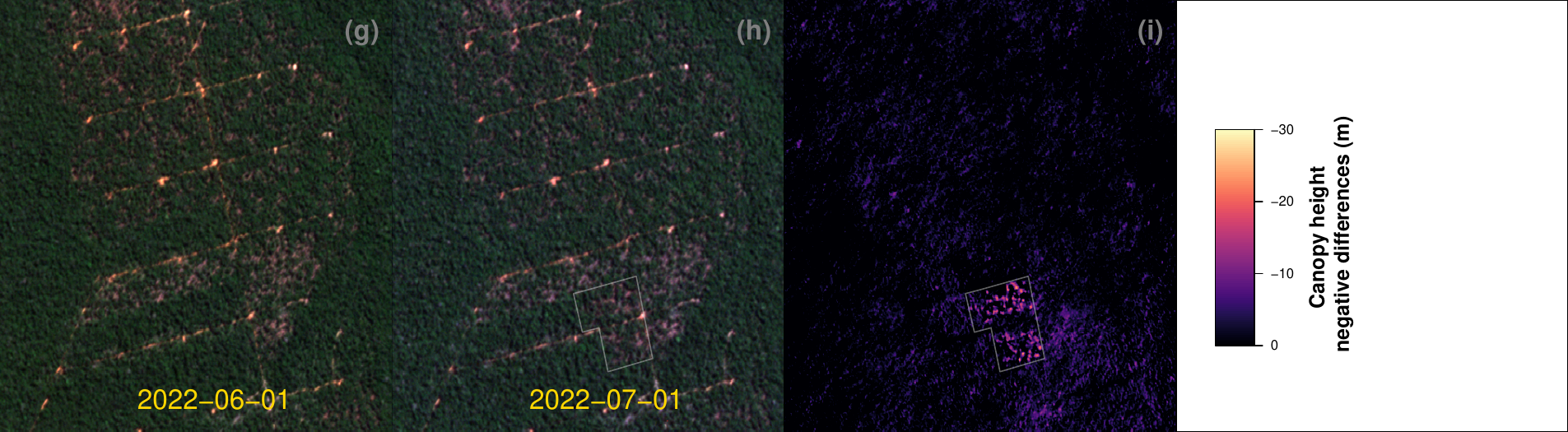}
 
 \caption{ Examples of height changes due to selective logging activities for the Planet quad 0714-0954 for the periods 2022-03-01 to 2022-04-01 (a-c), 2022-04-01 to 2022-05-01 (d-f), and 2022-06-01 to 2022-07-01 (g-i). Each row of images corresponds to one of these periods and includes: a color composite image of the area (left), a color composite image of the subsequent month showing logging activity (middle), and a canopy height difference map (right). The image size is approximately 2.45 km $\times$ 2.45 km. Image centroid in decimal latitude and longitude [-12.15929, -54.46710].}
\label{FigResLogging}
\end{figure}

In a Mato Grosso selective logging site (Fig. \ref{FigResLogging}), activities were tracked through 2022. The site was selected based on disturbance model results from \cite{Dalagnol2023} because it experienced successive logging events close in time, with minimal cloud cover, allowing clear visual assessment of these events. Between March and April (Fig. \ref{FigResLogging}a-b), new logging appeared below center and top-right, visible in RGB images and height difference maps despite illumination and geolocation noise (Fig. \ref{FigResLogging}c). From April to May (Fig. \ref{FigResLogging}d-e), logging moved above center, with clear canopy height reductions (Fig. \ref{FigResLogging}f). During June-July (Fig. \ref{FigResLogging}g-h), activity concentrated below center, with marked height reductions (Fig. \ref{FigResLogging}i). Despite some noise in the raw model output, large removed trees appear to show up in the canopy height differences between dates, suggesting the potential for detecting logging events at tree level.

\subsection{Clear-cut and regeneration detection from canopy height changes}

\begin{figure}[!ht]
\centering
\includegraphics[width=0.95\linewidth]{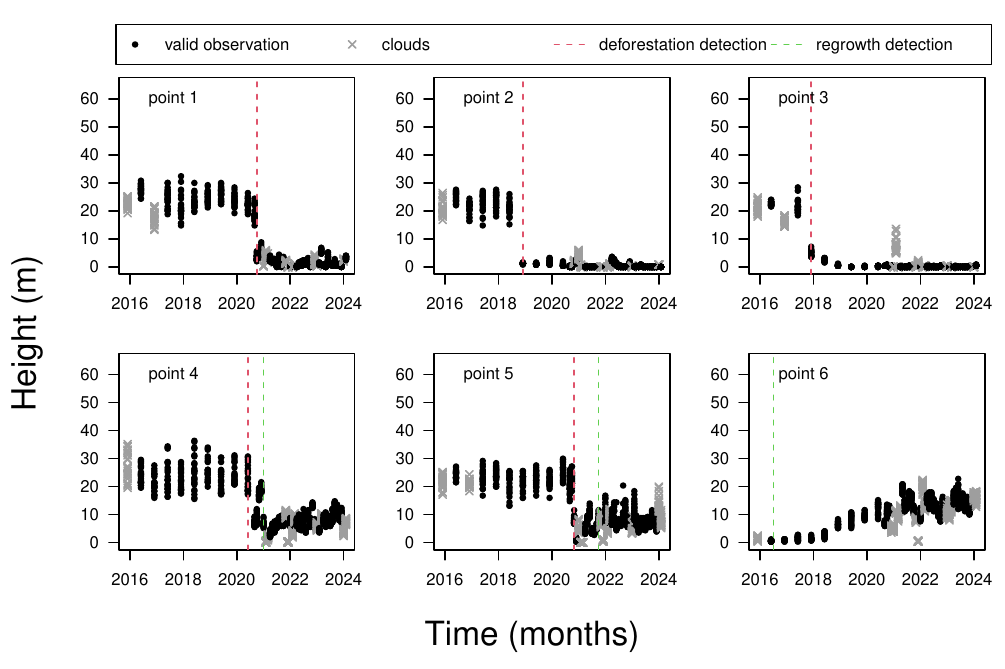}
\hspace{3.5cm}
\centering
\includegraphics[width=0.80\linewidth, trim={0cm 0cm 5.5cm 0cm},clip]{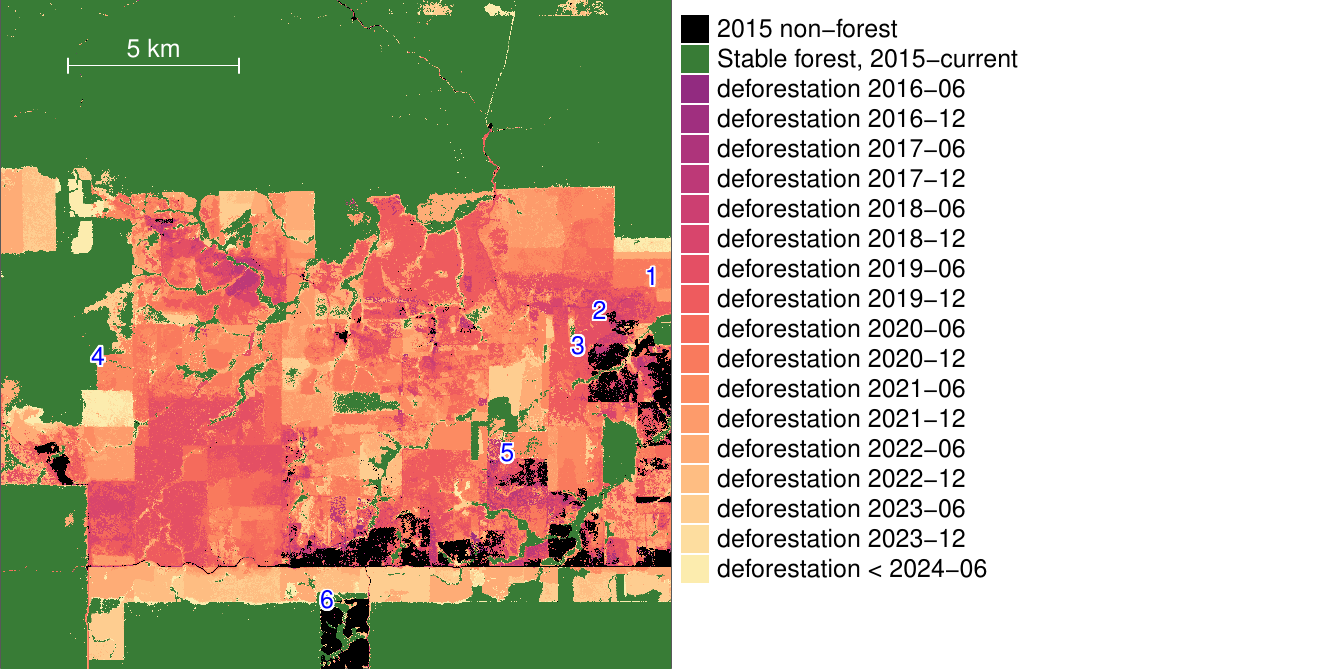}  
\caption{Example of height time series in case of deforestation (points 1, 2, and 3), deforestation and regrowth (points 4, 5) and only regrowth (points 6). For each location, the value of the point and its 24 closest neighbors are presented. To reduce the effect of possible geolocation errors between dates, we considered the neighboring pixels to assess the mean temporal behavior of an area rather than individual pixels. The locations of these points are displayed on our previously developed Planet NICFI-based deforestation product for the Planet quad 0681-0967 spanning from December 1, 2015, to February 1, 2024. The Planet NICFI quad covers approximately 19.5 $\times$ 19.5 km. Tile centroid in decimal latitude and longitude [-9.882275, -60.205078 ].}
\label{FigRestimeseries}
\end{figure}

Fig. \ref{FigRestimeseries} shows the height time series for deforestation and regrowth. Point 1 maintained stable height (2016-2019) before deforestation and has a sharp decrease in height in 2020, confirmed by our deforestation algorithm \citep{Wagner2023,CTreesREDD+AI2024}, with minimal subsequent regrowth. Point 2 showed stability until late 2018's deforestation, and Point 3's deforestation occurred mid-2017, both showing little regrowth. Points 4 and 5 demonstrate regrowth after deforestation in 2020, detected by our algorithm. Point 6, initially non-forest in 2015, shows continuous regrowth from 2018-2019, reaching 15-20 m by 2024. All time series show height variations due to cloud effects (Fig. \ref{FigRestimeseries}).

\subsection{Artifacts from clouds and shade}

\begin{figure}[!ht]
\centering
\hspace{2cm}
\includegraphics[width=0.90\linewidth, trim={0.1cm 0.1cm 5cm 0.1cm},clip]{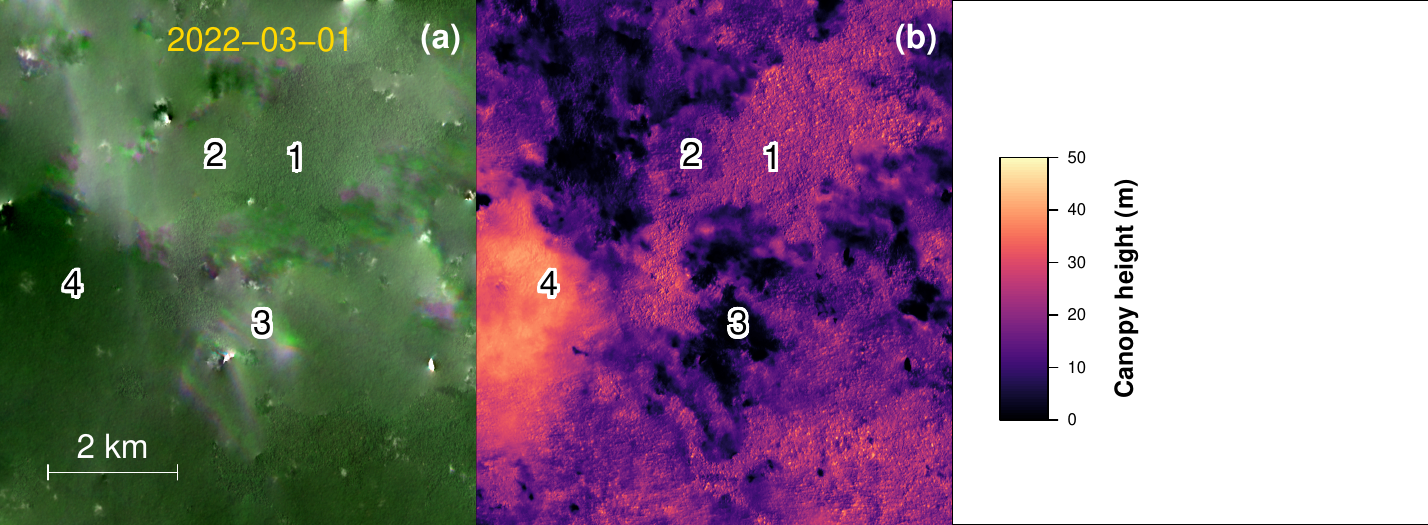}
 \caption{Example of artifacts in height prediction caused by haze, blur, clouds, and shade in the NICFI images before cloud correction for the Planet quad 0722-1038 basemap of 2022-03-01. Location 1 is a hazy but clear part of the image for reference, location 2 is a blurry part of the image, location 3 is a cloudy area with very high reflectance values, and location 4 is a shaded part of the image. Note that this image is classified as 100\% clouds after filtering by our cloud model. The image size is $\sim$ 7.34 $\times$ 7.34 km. Image centroid in decimal latitude and longitude [2.602864, -52.943115].} 
\label{FigResclouds}
\end{figure}

Raw predictions show various artifacts from clouds, haze, shade, and blur (Fig. \ref{FigResclouds}). The model handles light haze well, maintaining crown visibility (location 1, Fig. \ref{FigResclouds}a-b), but underestimates heights in blurred areas (location 2). Cloud presence reduces heights to zero, masking canopy features (location 3), while shaded forests show height overestimation (location 4). Our cloud mask enables us to discard all errors caused by clouds; for example, the entire image in Fig. \ref{FigResclouds}a is classified as cloudy.

\newpage

\subsection{Valid observations in Planet NICFI data} 
\label{cloudsvalid}

\begin{figure}[H]
\centering
\includegraphics[width=0.85\linewidth]{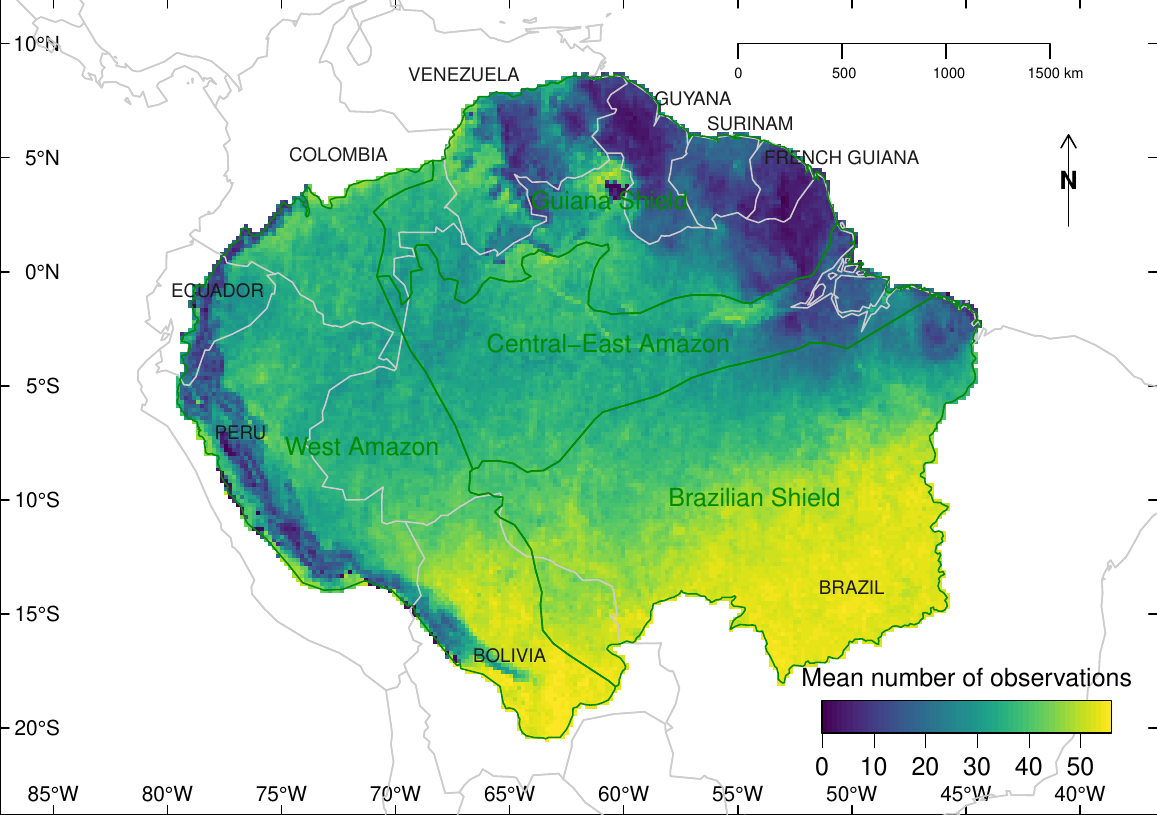} 
\caption{Mean number of cloud-free observations for the $\sim$ 20 km $\times$ 20 km pixels corresponding to the Planet NICFI tiles. Only pixels with vegetation height greater than zero were considered in the mean computation. The maximum number of observations is 56, corresponding to the 56 NICFI mosaic dates.} 
\label{FigGlobc}
\end{figure}

The Planet time series used in this study comprised 56 dates: 10 biannual basemaps (2015-12-01 to 2020-06-01) and 46 monthly basemaps (2020-09-01 to 2024-06-01). Thus, each pixel could have up to 56 valid (i.e., cloud-free) observations. In the following text, the number of observations is aggregated by NICFI tile to facilitate comparison between regions. Of 22,063 Amazon NICFI tiles, 8.06\% (1,860) had pixels lacking observations, though those tiles also had pixels with at least 3 observations. From 56 possible dates, tiles averaged 32.34 cloud-free observations, varying geographically (Fig. \ref{FigGlobc}). Southern Amazon showed highest coverage (up to 55.9 observations), while lowest counts occurred in French Guiana (0.25, quad \textit{0717-1048}), Guiana Shield, western Amazon near Andes, and central-western Amazon boundary. For low-observation regions, we made the cloud/shade masking less restrictive to increase valid observations (section \ref{masks}).

\newpage

\subsection{Amazon forest canopy height}
\label{AmzStat}

We found that the mean canopy height of the Amazon forest was 22.09 m, with a standard deviation of 5.270 m, a median of 22.25 m, and a 97.5th percentile of 32.10 m (Table \ref{tab1}).

\begin{table}[!ht]
\centering
\begin{tabular}{cc}
\hline
\textbf{Canopy height Metrics} & \textbf{value (m)} \\ \hline
Mean & 22.09 \\
Standard deviation & 5.270 \\
Percentile 2.5 &  11.34\\
Percentile 5 &  13.25\\
Percentile 25 & 18.75 \\
Median, Percentile 50 &  22.25\\
Percentile 75 & 25.54 \\
Percentile 95 & 30.54 \\
Percentile 97.5 & 32.10 \\ \hline
\end{tabular}
\caption{Amazon forest canopy height descriptive statistics.}
\label{tab1}
\end{table}

\begin{figure}[!ht]
\centering
\setlength{\unitlength}{1\textwidth}
\begin{picture}(0,0)
    \put(0.8,0.55){\textbf{(a)}}
    \put(0.8,-0.05){\textbf{(b)}}
\end{picture}
\includegraphics[width=0.85\linewidth]{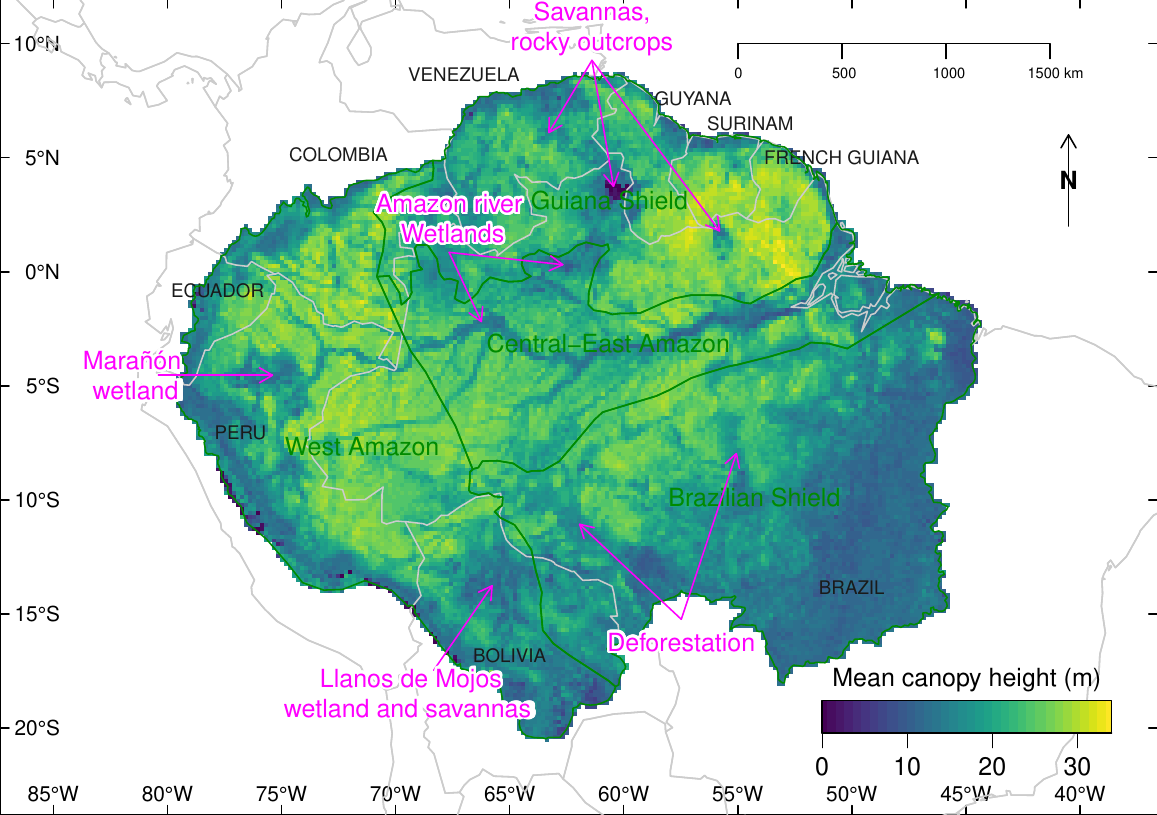} 
\includegraphics[width=0.85\linewidth]{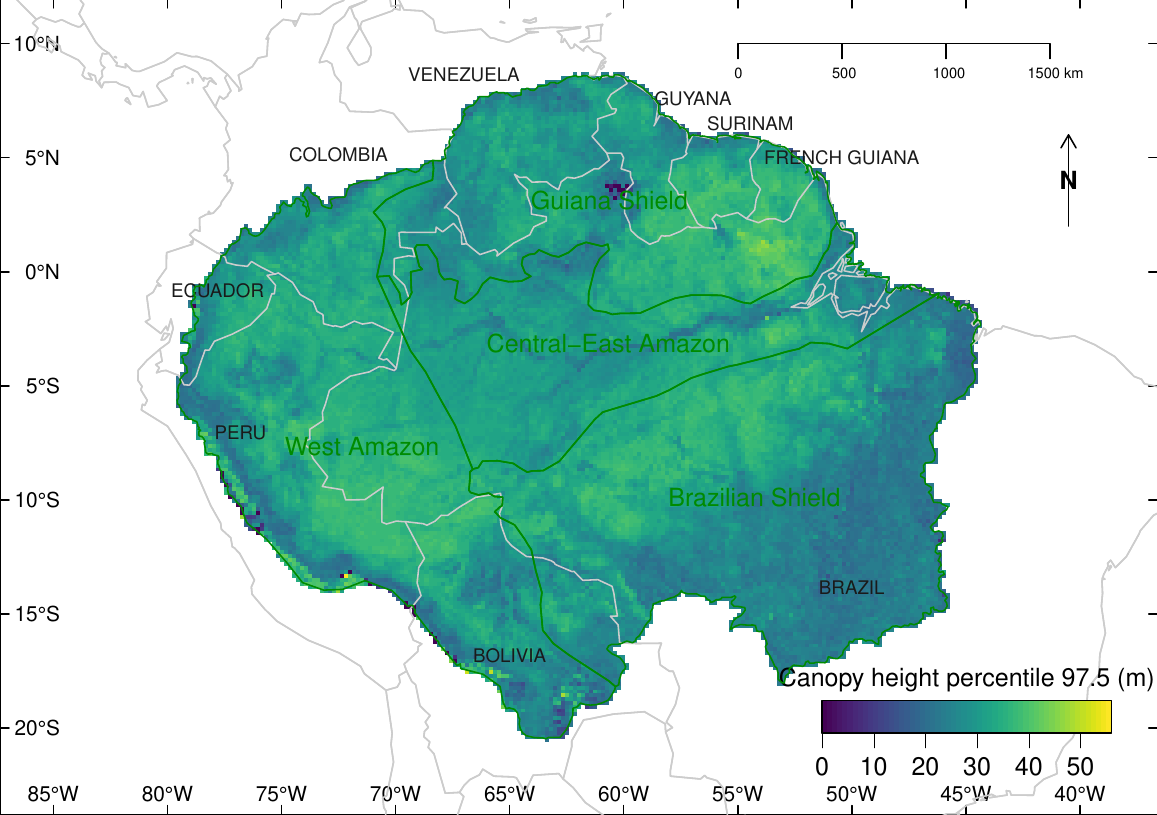} 
\caption{Map of the mean and percentile 97.5th of the Amazon forest tree canopy height on the period 2020-2024 (a-b). Each statistics is given for $\sim$ 20 km $\times$ 20 km pixels which correspond to a planet tile. Inside a tile, mean and percentile 97.5th are computed only with pixels that have $\geq$ three non-clouded observation.}
\label{FigGlob}
\end{figure}

Highest mean canopy heights (Fig. \ref{FigGlob}a) concentrate in the Eastern Guiana Shield and upper West Amazon/Central-East Amazon border, with Guiana Shield containing most continuous regions reaching $\sim$30 m.
The 97.5th percentile height distribution (Fig. \ref{FigGlob}b) forms a 1,000 km-wide circle around Central Amazon, exceeding 50 m in some areas. Notable tall forest hotspots include the French Guiana-Brazil border, areas south of the Amazon River ($\sim$500 km from Atlantic), and Western Amazon region.
On the western slopes of the Andes, there seem to be no natural forests, only tree plantations; the values (mean and 97.5th) represent only the few plantations within the 20 $\times$ 20 km tiles.

Lower canopy heights occur in several distinct regions: (i) near Amazon River and  wetlands, including Peru's Marañón Basin and Bolivia's Llanos de Mojos savannas; (ii) along Andean elevation gradients near treeline; (iii) around rocky outcrops, such as the Tepuis in Venezuela; (iv) in savanna ecosystems, with where the lowest canopy height at Brazil-Guyana-Venezuela borders; and (v) near deforested areas and particularly in the Arc of Deforestation.

\newpage

\subsection{Comparison with global canopy height products}
\label{CompaAmz}

\begin{figure}[!ht]
\centering
\includegraphics[width=0.40\linewidth]{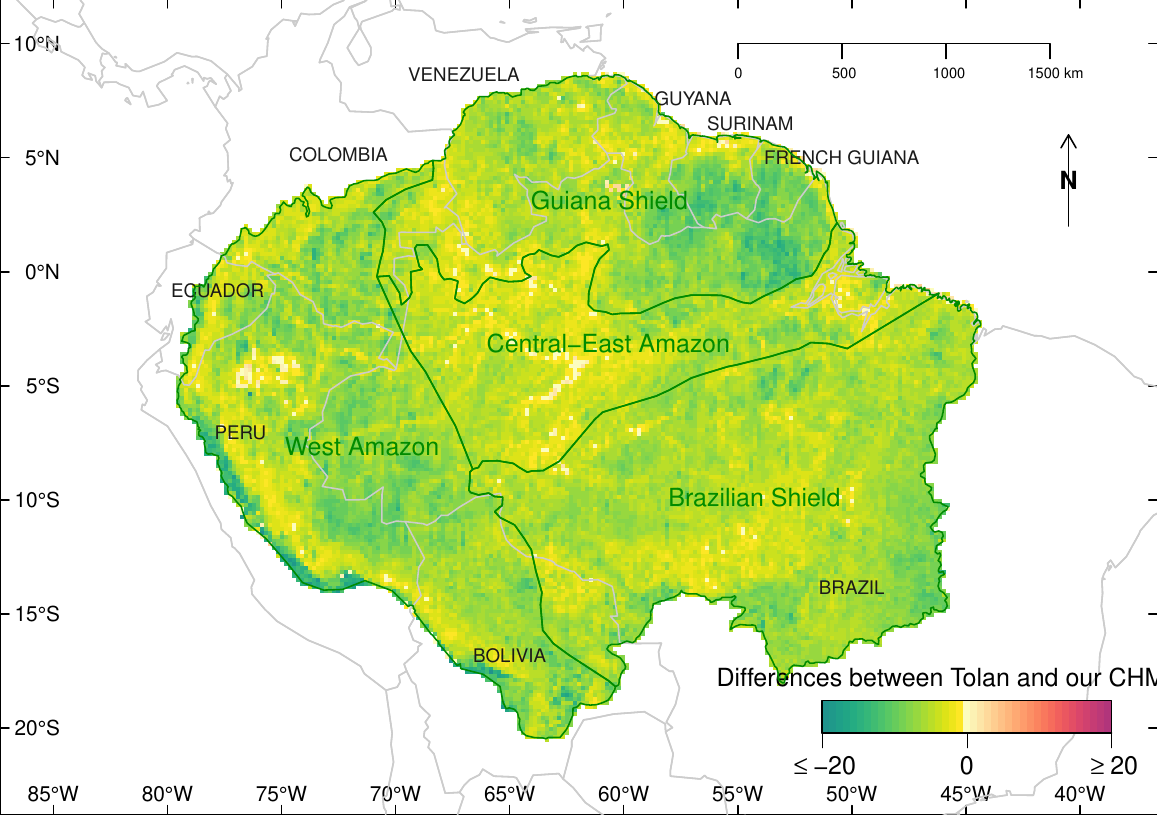} 
\includegraphics[width=0.40\linewidth]{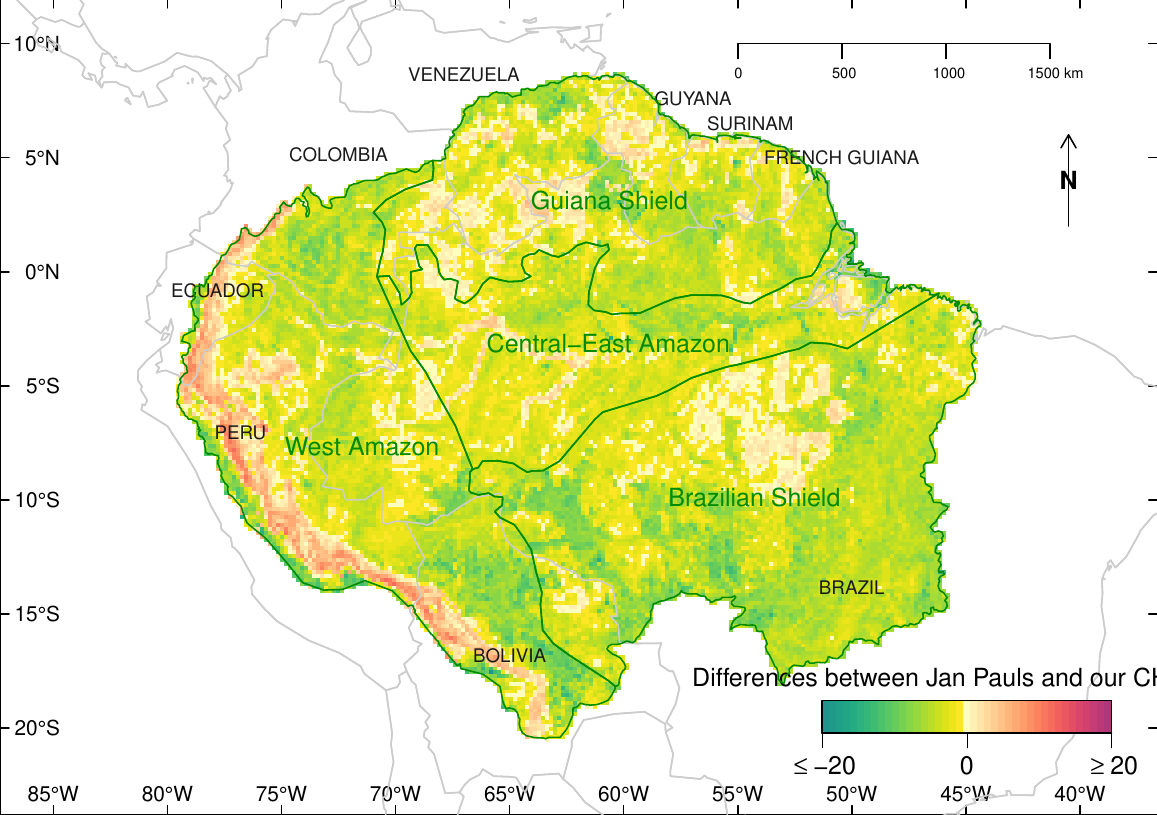} \\ 
\includegraphics[width=0.40\linewidth]{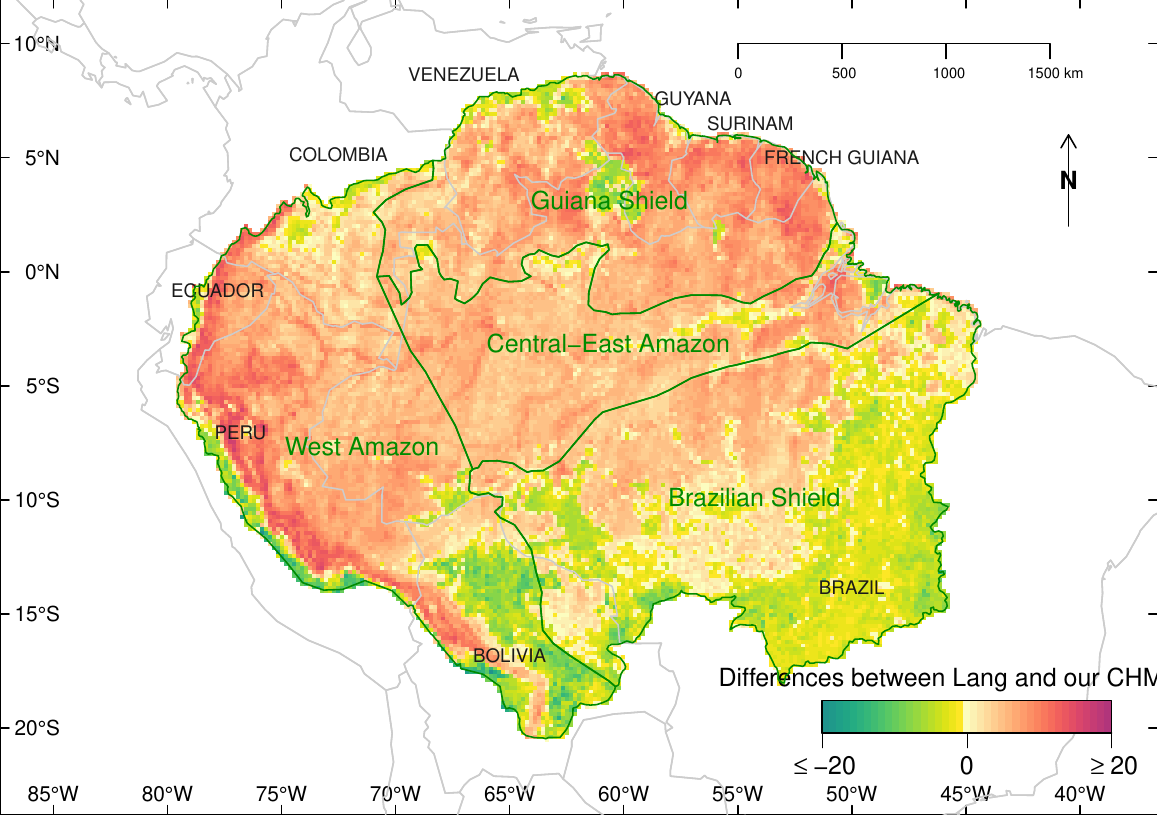} 
\includegraphics[width=0.40\linewidth]{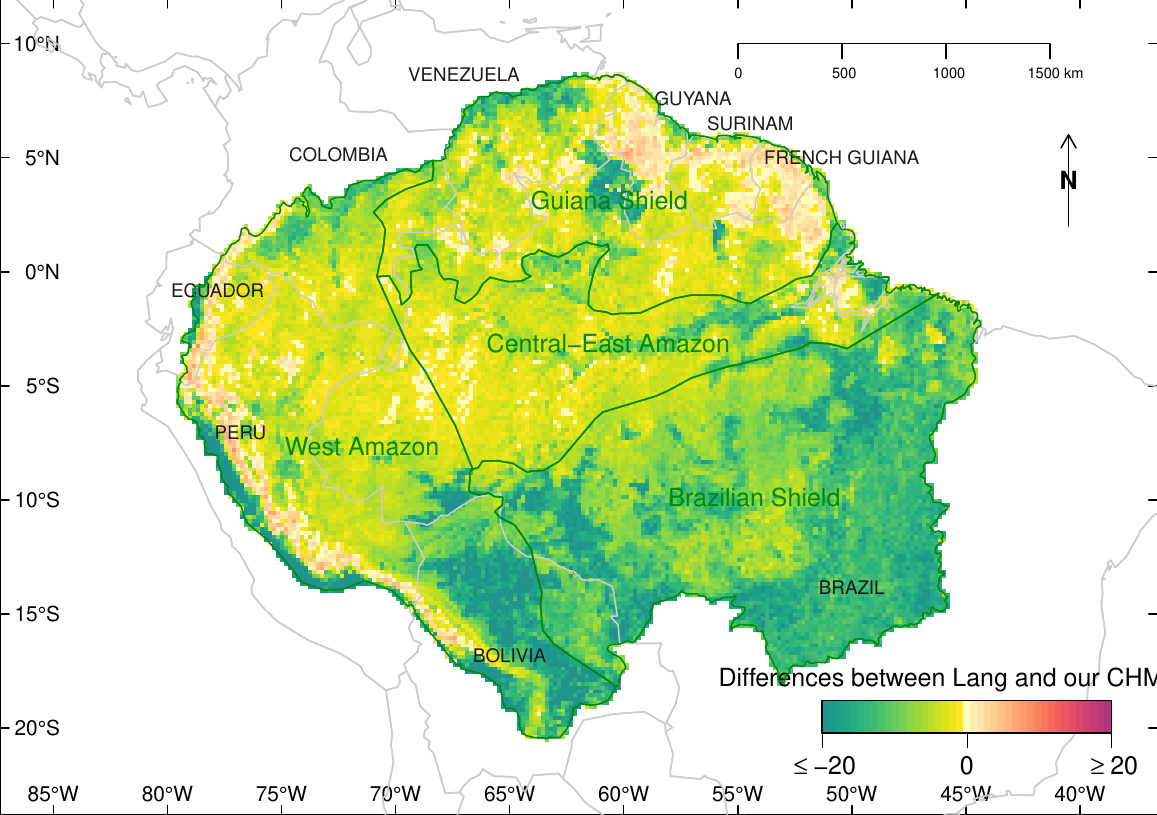} 
 \caption{Tree canopy height differences in meters between Tolan's product (a), Pauls's product (b), Lang's product (c-d), and our canopy height product. For Lang's product, we compared it to the mean of our product (c) and the 97.5th percentile (d), as Lang models the GEDI 98\% relative height (RH98). Differences are shown for the $\sim$ 20 km $\times$ 20 km Planet NICFI tiles, considering only pixels with vegetation height greater than zero. A negative difference means our canopy height is higher than the compared product.} 
 \label{FigDiffGlobc}
 \end{figure}

Compared to Tolan's model (Fig. \ref{FigDiffGlobc}a), our heights are higher ($>$10 m) in Guyana shield and Western Amazon, confirming Tolan's 30 m saturation (Fig. \ref{FigRes1}b), though both capture landscape height variations well. Pauls's model shows $<$10 m differences from ours (Fig. \ref{FigDiffGlobc}b), performing better near Andes due to combined radar-optical data \citep{Pauls2024}, but underestimating fragmented forests in the Arc of deforestation. Lang's mean RH98 exceeds our mean heights in dense canopy areas but is lower in sparse forests (Fig. \ref{FigDiffGlobc}c). Their RH98 versus our 97.5th percentile shows higher values in Guiana Shield and Andes, possibly due to better optical data coverage and geographic variables, but underestimates heights in fragmented forests (Fig. \ref{FigDiffGlobc}d).

Overall, our model provides higher height estimates across Amazon, except in high cloud/shade regions near Andes, while maintaining good spatial agreement with other models despite different data sources.

\newpage

\subsection{High resolution map of the Amazon canopy height}

\begin{figure}[ht]
\centering
\includegraphics[width=0.85\linewidth]{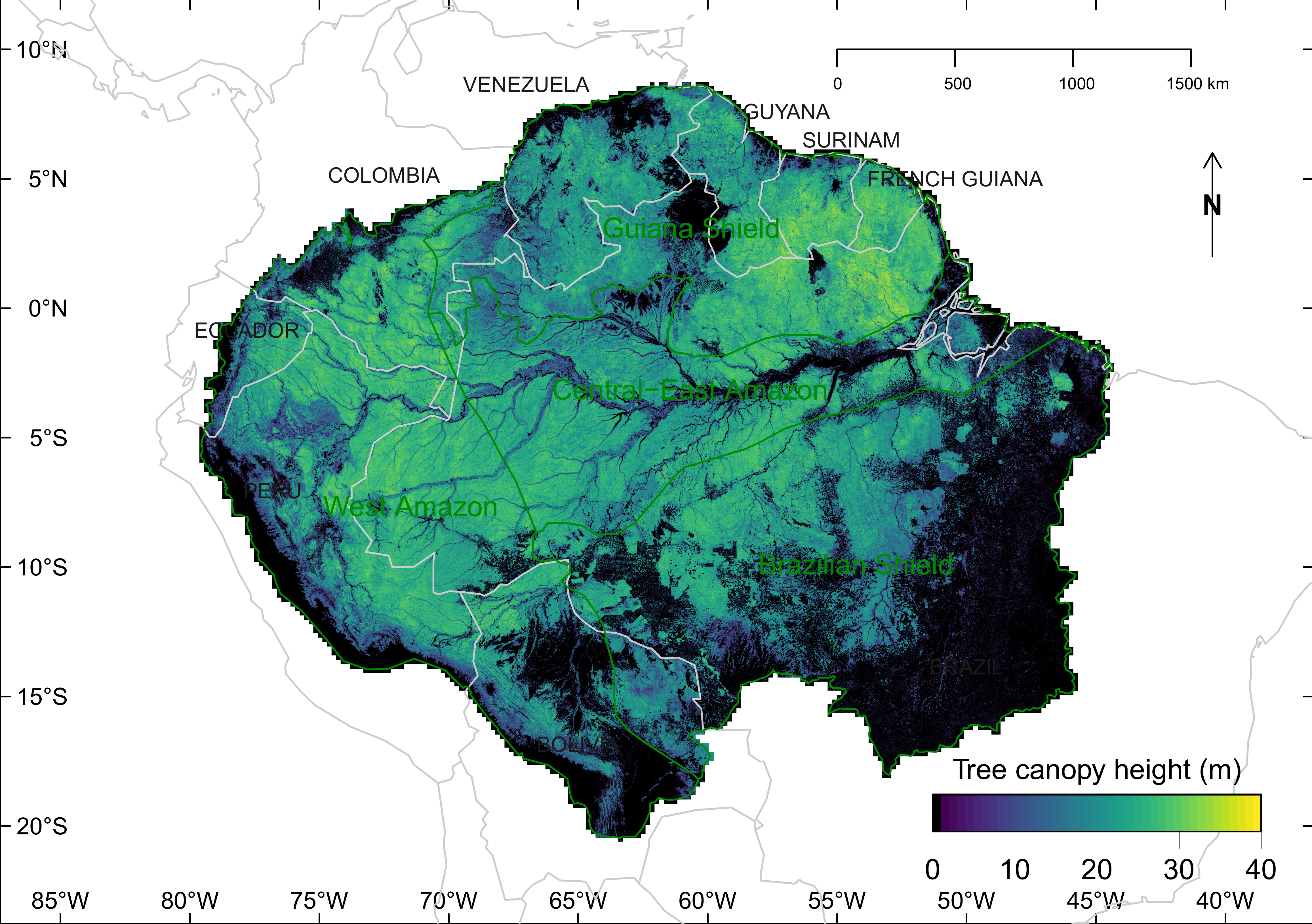} 
\caption{Canopy height of the Amazon forest (m). To facilitate visualization at very high resolution, the colors represent the estimates from our model, aggregated to an 80 m spatial resolution using the median.}
 \label{FigGlobVHR}
 \end{figure}

The very high-resolution canopy height map of the Amazon forest shows highly diverse and consistent large-scale spatial patterns of tree height (Fig. \ref{FigGlobVHR}). As observed in the analysis (section \ref{AmzStat}), variables associated with lower heights include (i) natural factors such as wetlands, savannas, and rocky outcrops, and (ii) human activities like deforestation and roads. In Brazil, roads are already following the arc of taller forests revealed by the p97.5 canopy height (Fig. \ref{FigGlob}b), and some large forested areas within this arc have already been deforested and fragmented.

\begin{figure}[H]
\centering
\includegraphics[width=0.85\linewidth]{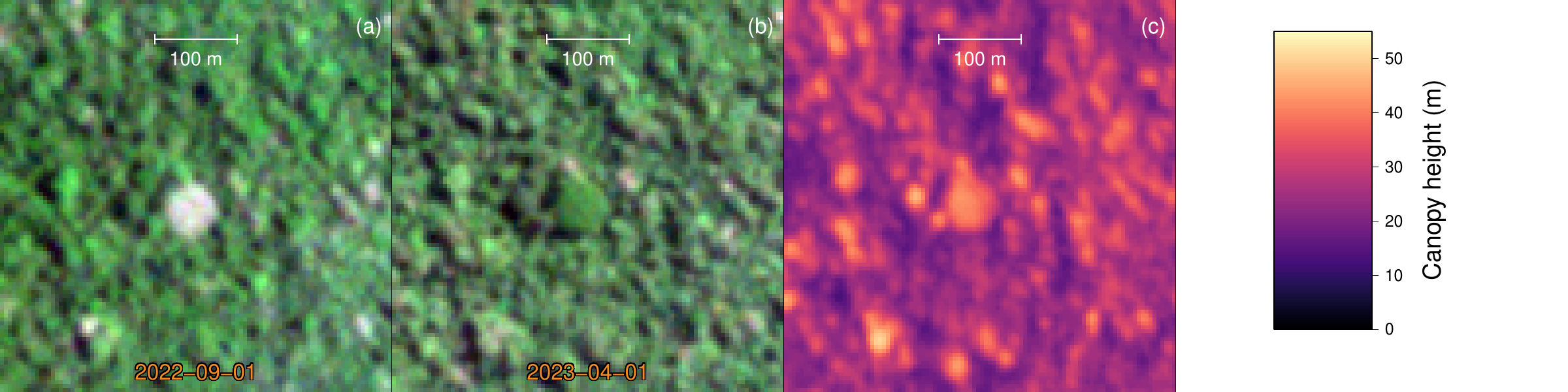} 
\includegraphics[width=0.85\linewidth]{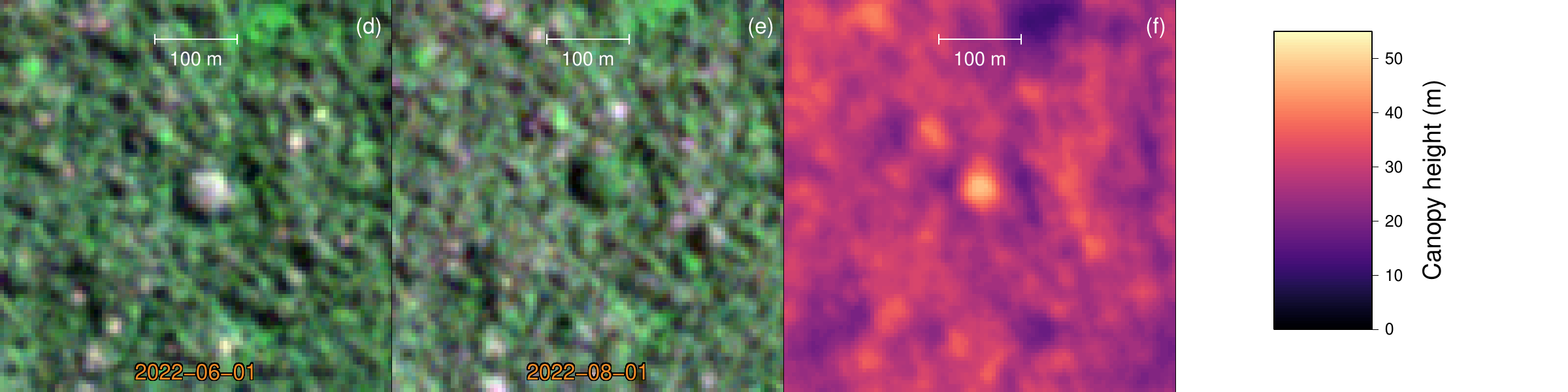} 
\includegraphics[width=0.85\linewidth]{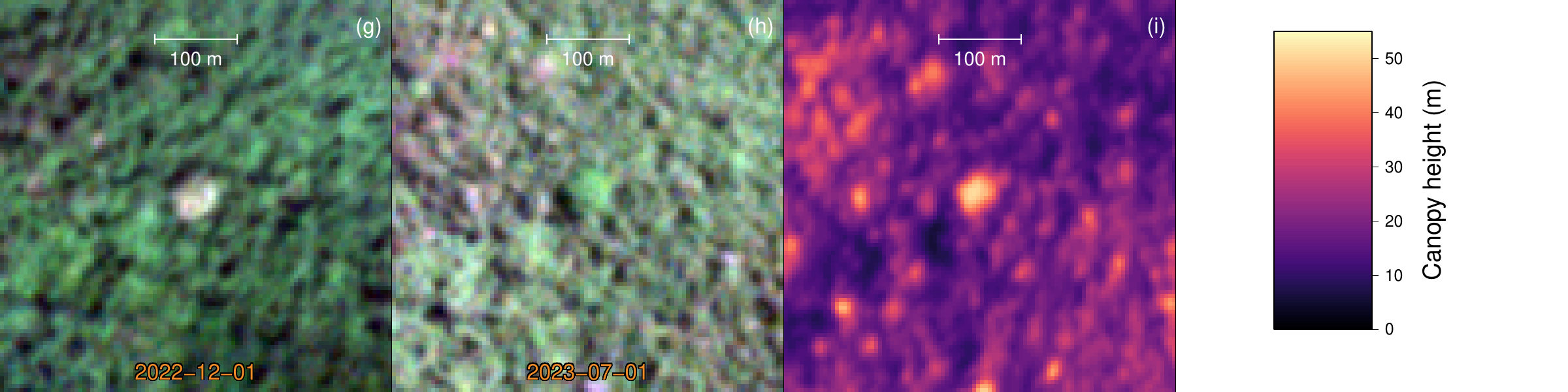}
\includegraphics[width=0.85\linewidth]{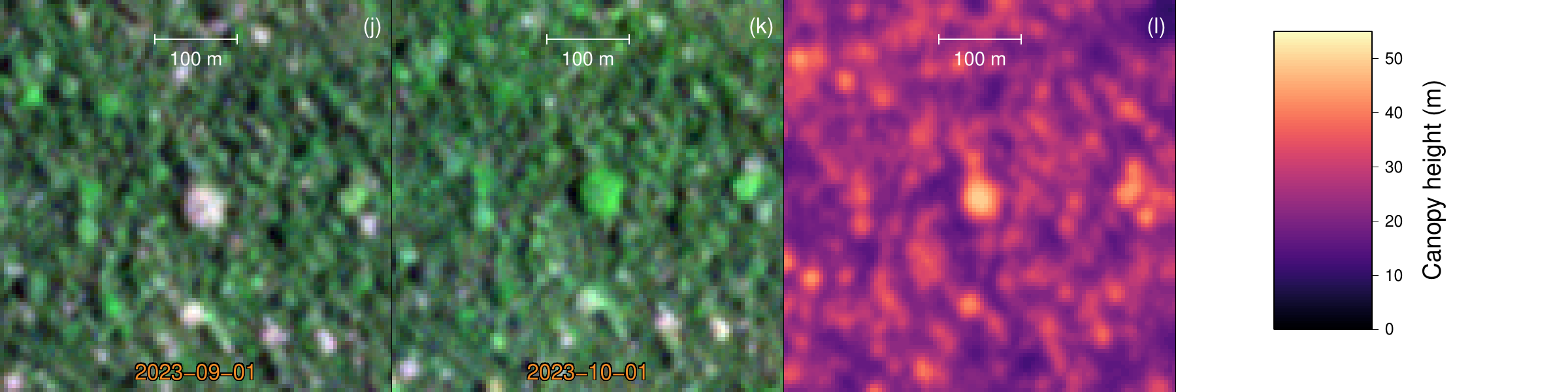}
\caption{Examples of giant trees identified by our Amazon canopy height model. For each individual tree (row), a Planet NICFI image is shown during its leafless period (first column), during its leaf-covered period (second column), and the third column presents our canopy height composite for the year 2020. The resolution of the NICFI image, preserved in this figure, is 4.78 m. Tree coordinates in decimal latitude and longitude [-10.6957146, -67.03241];
[0.2214842, -75.18744];
[6.4737808, -61.35848]; and
[-7.6873982, -68.93119].}
\label{FigGiants}
\end{figure}

\newpage
For most of the Amazon forest, the model enables the detection of the tallest trees with the largest crowns, Fig. \ref{FigGiants}. These trees were visually identified by examining groups of pixels above 40 meters across some regions of the Amazon. The first individual, Fig. \ref{FigGiants}a-c, has a crown diameter of approximately 70 m, making it the largest tree found in our Amazon dataset so far, and likely one of the largest tree crowns in the Amazon forest. All the other trees also have crown diameters above 50 meters, which explains their good visibility in the NICFI image with a spatial resolution of 4.78 m. Each of these trees exhibits distinct annual phenology, with a complete renewal of leaves occurring within 1-2 months every year during the same period (observed from 2021-2024). The species are unknown.

\newpage
\section{Discussion}

Here, we show that Planet NICFI images with a 4.78 m spatial resolution enable accurate mapping of Amazon forest canopy height, even allowing the identification of most large tree crowns. The U-Net network, adapted for regression, estimated canopy height with a mean error of 3.68 m on the validation dataset. This confirms the high capacity of convolutional networks to support vegetation mapping, particularly canopy height mapping with high spatial resolution images \citep{Wagner2024, KATTENBORN202124}. However, since the discontinuation of the Planet NICFI program, both historical and real-time Planet imagery are no longer open access. This will limit the application of the method for academic, non-governmental, and governmental organizations due to the cost of commercial data.

We found that the Amazon forest has a mean canopy height of 22.09 m, a median of 22.25 m, and a 97.5th percentile of 32.10 m. The tallest forests form a 1,000 km-wide arc around the Central Amazon. A canopy height hotspot was identified in the Guiana Shield, near the southern border of French Guiana and Brazil, where the tallest documented Amazon trees, such as a \textit{Dinizia excelsa} reaching 88.5 m, have been recorded \citep{Gorgens2019}. On the Amazon forest scale, the canopy height varies significantly, with shorter forests near rivers, wetlands, savannas, and deforested areas, reflecting local environmental and anthropogenic influences (Fig. \ref{FigGlobVHR}).

For the Amazon forest, our locally calibrated model provided more accurate canopy height estimates, with a better prediction range and less saturation, than available global models (Fig. \ref{FigRes1}, \ref{FigRes5}). It is the only one that effectively captures tree heights beyond 45 m. While expected for lower-resolution models \citep{lang2022, Pauls2024}, it also outperformed Tolan's higher-resolution (0.5m) Maxar-based model \citep{tolan2023sub}, suggesting high-quality local training data's importance over resolution \citep{Rolf2024}. Several factors could explain why our height estimates are better. First, the U-Net architecture uses all local information with only minimal compression, whereas other architectures, such as Dino V1 used in Tolan's model, seem to lose fine-scale details due to compression. Second, the $\sim$5 m resolution of our CHM dataset used for training produces smoother crowns, likely making predictions easier than for 1 m CHMs, which may contain significant intra-crown variation, while at larger scales, such as 10 m or 30 m, individual crowns are no longer visible. Finally, the extensive aerial LiDAR training data covering a wide range of heights across Brazilian tropical forests likely contributed, while Tolan's model used only a small portion from the Atlantic Forest, which lacks tall trees. Our canopy height map, Fig. \ref{FigGlobVHR}, shows similar spatial patterns to other maps \citep{lang2022, tolan2023sub, Pauls2024}, but reaches greater heights and provides crown-level detail, enabling temporal monitoring.

Some regions exhibit higher canopy heights, such as the Guiana Shield and the western Amazon (Fig. \ref{FigGlob}b), but the most prominent patterns of canopy height seem to be influenced by edaphic conditions. Studies in the Amazon have reported that forests on plateaus with well-drained clay soils tend to have taller and more uneven canopies compared with forests in areas along rivers, on white-sand soils, or on shallow soils \citep{Rosa2017, Goncalves2024, Laurance1999, Quesada2012, Feldpausch2011}. In other words, poorly drained, shallow, or rapidly draining soils tend to support lower biomass and shorter trees. Consistent with these previous findings, our map shows the lowest canopy heights   near wetlands, major rivers, high elevations, rocky outcrops, and savannas, such as those in southern Guyana and the Roraima region (Fig. \ref{FigGlob}a, Fig. \ref{FigGlobVHR}). Maximum height appears to represent an upper limit achievable by most Amazon forests, as shown by the 97.5th percentile in Fig.~\ref{FigGlob}a, regardless of forest basal area or mean wood density, which are known to increase from the driest to the wettest regions and from the west to the east, respectively \citep{Malhi2006, Avitabile2016}. Finally, our map highlights large-scale human disturbance effects on Amazon forest height, with visible scars of deforestation in the Brazilian Shield and western Amazon region (Fig. \ref{FigGlob}a, Fig. \ref{FigGlobVHR}).

The Amazon forest has high spatial variability in species \citep{Luize2024}, forest structure, and biomass \citep{Conto2024, Saatchi2011}. Our large-scale height patterns reinforce that the Amazon is a highly heterogeneous forest, with notable features like the tall tree hotspot in the Guiana Shield \citep{Gorgens2019}. This heterogeneity raises particular conservation concerns, as deforestation in the Amazon still targets large continuous forest patches, Fig. \ref{FigGlobVHR}, and removes unique forest ecosystems.

Our map may underestimate Andean canopy heights due to cloud cover and shading, where radar-based models like Pauls' are more reliable \citep{Pauls2024, lang2022}. While we used strict cloud masking with 1200 m buffers to ensure very high-quality, cloud-free pixels, but this significantly reduced usable pixels in cloudy regions. Future model iterations will aim to predict heights under thin clouds to improve coverage and accuracy in Andean slopes. We also acknowledge that nearly all predictions outside Brazil are out-of-domain extrapolations, and independent assessments of CHM quality in these regions will be made as new LiDAR datasets emerge and are encouraged for those who have access to such data.

While rare Amazon trees can reach 65-88.5 m \citep{Gorgens2019}, our sample shows $\sim$ 20 m mean canopy height, with 22.1 m Amazon-wide average. GEDI data shows canopy heights $>$40 m represent 5\% in Amazon, 10\% in African, and $>$25\% in Asian rainforests \citep{Bourgoin2024}. Southeast Asian \textit{Dipterocarp} forests are tallest, with mean heights recorded for a forest of Borneo averaging 50 m with individuals up to 98 m \citep{Shenkin2019}.  We acknowledge that our model cannot currently capture these extremely tall trees and future work will incorporate other tropical forests to improve tall tree height estimation.

The intact Amazon offers a unique opportunity to study pristine tropical forests at scale. Our model could be used to locate all the giant trees in the Amazon forest (Fig. \ref{FigGiants}), which are crucial for biodiversity and ecosystem function \citep{piirto2002ecological, Francis2019, enquist2020}, to plan their protection against near-future risks. The high-resolution canopy height maps produced in this study could also be used as input for models estimating forest carbon stocks, as tree height is linked to aboveground biomass and carbon storage \citep{asner2014m, Saatchi2011}. Our height map also provides unprecedented canopy structure data where LiDAR is unavailable, which could inform on biodiversity to some extent, as forest structure and canopy height are indicators of habitat complexity and can be associated with species diversity \citep{Roll2015, Martins2017, enquist2020, Moudry2024}. Furthermore, our map could serve as a benchmark for monitoring forest degradation in the future, enabling the detection of height changes within the canopy at the tree level (Fig.~\ref{FigResLogging}) that might not be apparent in lower-resolution height maps based on Landsat or Sentinel images.

\subsection{Advances in canopy height changes mapping}

Currently, tropical forest cover changes from deforestation, degradation, or regeneration at regional to global scale are mostly made based on the semantic segmentation and time series analysis of classified pixels \citep{Hansen850,MapBiomas2024,Wagner2023, Dalagnol2023,csillik2019,heinrich2021,Heinrich2023}. In regrowth analysis, the algorithm relies on forest cover classification rather than canopy height. For example, if a pasture becomes classified as forest and remains so for 10 years, the pixel is considered a 10-year regenerating secondary forest \citep{heinrich2021}. This works well for regrowth after large tree cover removal or initially non-forested pixels, but logged or degraded pixels can lose height while remaining classified as unchanged forest cover. The canopy height could help in this case, but the remote sensing community is just beginning to map tropical forest cover changes using height. Most recent works on multi-temporal height estimation use LiDAR data because open global canopy height products only contain a single date \citep{lang2022, Pauls2024, tolan2023sub, potapov2021m}. In the Amazon arc of deforestation, repeated LiDAR measurements demonstrated that forest height losses, carbon changes, and growth from both anthropogenic and natural processes could be identified and estimated \citep{Csillik2024}. Similar datasets revealed dynamics of gap formation and closure, including tree growth in gaps \citep{dalagnol2021, Winstanley2024}. Currently, only Planet's commercial Forest Carbon Monitoring (FCM) product provides global quarterly updates at 3 m resolution for canopy cover, height, and aboveground carbon, though it remains non-peer-reviewed.

In this work, we demonstrate Planet NICFI time series' potential for mapping high-resolution canopy height changes, Fig. \ref{FigResLogging} and \ref{FigRestimeseries}. Our model captures logging activities and their impact on forest canopy height, Fig. \ref{FigResLogging}. Despite some noise and phenology-related changes outside logging areas, detecting height decreases is promising, as such disturbances are extremely difficult to estimate from optical remote sensing. While locating activities as selective logging or fire impacts is currently possible \citep{Dalagnol2023}, quantifying height and biomass changes remains challenging. High-frequency time data, such as monthly Planet NICFI are crucial for tree loss detection, as tropical forest gaps can close within months through lateral ingrowth, potentially masking tree loss \citep{dalagnol2021, Winstanley2024}. Using height time series (Fig. \ref{FigRestimeseries}), deforestation appears as height changes, matching previously identified deforestation dates from tree cover changes \citep{Wagner2023}. Forest-to-pasture conversion shows near-zero heights (Fig. \ref{FigRestimeseries} points 1-3), while regrowth cases show increasing heights (Fig. \ref{FigRestimeseries} points 4-5). Though growth detection requires years, secondary forest growth detection is promising. The height time series (Fig. \ref{FigRestimeseries}, point 6) enabled tracking changes in tree height for a forest regrowing from abandoned pasture since 2016. While requiring further validation, these results suggest potential for measuring regrowth using open-access NICFI images, previously achievable only through repeated LiDAR or commercial data.

\subsection{Implications for Ecology and Conservation}

The Amazon forest, as one of the largest carbon pools in terrestrial ecosystems, is a key element of the global carbon cycle and climate regulation \citep{Mo2023,Gatti2021,Pan2024,Rosan2024}, and mapping its biomass remains a major challenge. Our map alone cannot be directly converted into biomass estimates, as additional parameters on the vertical distribution of vegetation would be needed \citep{Longo2016}. However, it can provide information for classifying forests by height and structural attributes, which may serve as proxies for biomass. As our map provides the finest detail of canopy height over the Amazon so far, it becomes possible to identify forests based on their canopy height characteristics and target them for sampling of biodiversity and biomass. In some cases, the canopy height characteristics alone may also justify considering certain forests for protection, for example the Amazon tallest forests in the Amapá region of Brazil.

Our model is the first that can estimate heights above 45 m in the Amazon forest, Fig. \ref{FigRes1}. Furthermore, the largest individual trees are clearly visible in the predicted CHM, Fig. \ref{FigRes2a}, \ref{FigRes2b} and \ref{FigGiants}. While the  tree crowns can be observed in RGB images, they are further enhanced in the canopy height visualizations, allowing the larger and tallest individuals to be visually identified, Fig. \ref{FigGiants}. For example, visually locating the giant trees, each located in a different country (Fig. \ref{FigGiants}), required less than two hours. Our map could be used to locate the largest individuals and the tallest forest of the Amazon forest which is a primary conservation importance, as large trees, being part of the megabiota, are more susceptible to extinction, and changes in their abundance disproportionately affect ecosystem and Earth system processes, including biomass, carbon, nutrients, and fertility \citep{enquist2020}.

In the Amazon, managed forest use mainly involves selective logging, which does not remove the entire canopy cover but only selected stems, and illegal logging (aside from clear cut) tends to follow the same practice, where only a few stems are removed while the forest cover remains. Currently there is no satellite-based data sufficiently accurate or with sufficient spatial coverage to detect the (illegal or legal) removal of individual trees. It is only currently made using Lidar data \citep{Csillik2024, dalagnol2021, Longo2016} or VHR images \citep{dalagnol2019quantifying}. Here, we show that just by taking the raw difference in predicted height between two dates, the location removed individual crowns could be observed, Fig. \ref{FigResLogging}, which suggests that the detection of individual tree removal by logging is feasible. In further work, we will analyze the changes detected by our approach, which will require validation with LiDAR data and temporal filtering or annual aggregation to remove noise from illumination, phenology and geolocation errors.

The restoration of tropical forests is considered one of the most effective strategies for climate change mitigation \citep{Bastin2019,Hasler2024}. However, there is currently no widely established model to monitor forest regrowth, and remote sensing methods for this purpose are still under development. To our knowledge, only two studies have applied deep learning to measure forest regrowth \citep{Schwartz2025,Wagner2025}. While not validated with LiDAR, we show here that our algorithm was able to detect growth in height in a regenerating forest, Fig. \ref{FigRestimeseries}. In a recent study motivated by this observation, we applied the model developed in this work to $Pinus$ and $Eucalyptus$ plantations from 0 to 8 years in Southwest Brazil, where we also had LiDAR measurements \citep{Wagner2025} and found that it can accurately track the regrowth of $Pinus$ and, to a lesser extent, $Eucalyptus$, but yielding promising results for natural forest monitoring. In future work, we plan to explore and validate the application of the model to tropical forest regrowth, aiming to develop robust methods for large-scale monitoring of forest restoration.

Finally, The Amazon forest structure is diverse, and the LiDAR sample currently available does not yet capture the full range of ecosystems, such as the bamboo-dominated forests in Acre, the giant tree forests in Amapá, or areas with fertile soils influenced by the Andes, nor the large-scale nutrient and elevation gradients present across the basin \citep{Dalagnol2018,Gorgens2019,Gorgens2021, Malhi2006, Quesada2012, Asner2014}. Future work could examine whether these ecosystems or gradients are detectable in our map to further improve knowledge of domain-scale variation in height and biomass, and, if not, investigate the underlying reasons to guide further improvements of the map.

\subsection{Open data of canopy height}

Our dataset is the first to enable the visualization of all the largest tree crowns in the Amazon. This raises questions regarding open data, as sharing such datasets could impact the surrounding communities and environments \citep{Bennett2024}. These large trees may be located on indigenous lands, areas with uncontacted people, or public/private lands. Public knowledge of their locations could attract scientific research, tourism, or logging, with unclear benefits or harms. While many large trees remain protected by isolation in remote areas, others near deforestation may only be known to loggers. Logging industries can access this information through aerial surveys or field exploration, but these trees remain unknown to local communities, governments, conservationists, and tourism operators who could protect and sustainably use them. These trees could be fully protected or, if used in sustainable eco-tourism, could contribute to socio-bioeconomies (SBEs) \citep{Garrett2024}, defined as economies based on the sustainable use and restoration of Amazonian ecosystems, as well as Indigenous and rural livelihoods in the region.

A potential solution could involve developing a monitoring system to raise awareness about these trees, thereby increasing their protection by making their existence more visible to actors other than the loggers. As more people become aware, these trees, rather than being at greater risk, could actually benefit from the attention and protection brought about by public visibility, in an analogy to the use of public space in cities \citep{Jacobs1961}. Following the demand of the Remote Sensing community and experts, our dataset have been made available on  \href{https://registry.opendata.aws/ctrees-amazon-canopy-height/}{AWS Open Data Registry}, and in the \href{https://gee-community-catalog.org/projects/ctrees_amazon/}{Awesome Google Earth Engine (GEE) Community Catalog}.

\section{Conclusion}

We present the first complete Amazon canopy height map at 4.78 m resolution (2020–2024), generated using a U-Net model trained on Planet NICFI imagery and airborne LiDAR data. We demonstrate that a U-Net model trained locally with high-quality LiDAR–Planet pairs provides reliable canopy height estimates at fine spatial resolution, with our model outperforming existing global products from Sentinel-2/1 and Maxar Vivid2. Our model is the only one able to capture heights above 45 m, and the resulting Amazon map captures key structural patterns, from the tallest trees in the Guiana Shield to shorter vegetation along wetlands, rivers, rocky outcrops, savannas, and high elevations. Finally, we show that this resolution enables, to a certain extent, the detection of small-scale canopy height changes associated with deforestation, logging, and regeneration, making it a potential new resource for monitoring forest structure across the Amazon.

\section{Acknowledgments} 
The authors wish to thank the Grantham Foundation and High Tide Foundation for their generous gift to UCLA and support to \url{CTrees.org}. Part of this work was carried out at the Jet Propulsion Laboratory, California Institute of Technology, under a contract with the National Aeronautics and Space Administration (NASA).
Conceptualization, F.H.W., R.D., G.C., S.F. and S.S.; methodology, F.H.W., R.D, G.C., M.C.M.H., S.G. and S.F.; software, F.H.W., S.G., R.D., G.C., S.F. and M.C.M.H.; validation, F.H.W. and L.B.S.T. ; formal analysis, F.H.W. and S.G.; investigation, F.H.W., R.D., G.C., S.G., L.B.S.T., S.F., M.C.M.H. and S.S.; resources, S.S.; data curation, F.H.W., M.C.M.H., S.G., L.B.S.T., M.K. and J.P.H.B.O; writing—original draft preparation, F.H.W.; writing—review and editing, F.H.W, R.D., G.C., M.C.M.H., S.G., L.B.S.T., S.F., M.K., J.P.H.B.O, L.A., C.C., S.P.G-C., S.L., Z.L.,  A.M., Y.Y., E.G.S., S.R.W., M.B., P.C., S.C.H. and S.S.; visualization, F.H.W. and M.C.M.H.; supervision, S.S.; project administration, S.C.H. and S.S.; funding acquisition, S.C.H. and S.S.

\section{Data availability} 
Our Amazon canopy height dataset is available on  \href{https://registry.opendata.aws/ctrees-amazon-canopy-height/}{AWS Open Data Registry}, and in the \href{https://gee-community-catalog.org/projects/ctrees_amazon/}{Awesome Google Earth Engine (GEE) Community Catalog}. At the time of the study, Planetscope imagery in tropical areas via Norway’s International Climate and Forest Initiative (NICFI) was available for non-commercial purposes from Planet Labs. They are now only available commercially from the Planet tropical forest observatory \url{https://www.planet.com/tropical-forest-observatory/}. The LiDAR datasets from the “Sustainable Landscapes - Brazil”  publicly available 
 at \url{https://daac.ornl.gov/CMS/guides/LiDAR_Forest_Inventory_Brazil.html} and \url{https://www.paisagenslidar.cnptia.embrapa.br/}. Data and metadata of the EBA (Biomass Estimation of the Amazon) project LiDAR transects across the Brazilian Amazon are publicly available from \url{https://zenodo.org/records/7636454} and \url{https://zenodo.org/records/4968706}. The LiDAR dataset covering the São Paulo Metropolitan Region (SPMR) are publicly available and distributed in LAZ files at \url{https://registry.opendata.aws/pmsp-lidar/}.

\newcommand{\newblock}{}
\bibliographystyle{elsarticle-harv}
\bibliography{references_height_tropical} 








\end{document}